# A bilingual AI audiologist built through rubric-guided closed-loop engineering outperforms human audiologists in a blinded evaluation of simulated cases

Linkai Li[1,2*], Changgeng Mo[2*], Hanlin Yu[1,2], Congxi Lu[2], Shangqiguo Wang[3], Matthew B Fitzgerald[4], Shan X Wang[1†]

[1]Department of Electrical Engineering, Stanford University, Stanford, CA, USA

[2]Orka Labs Inc., Shanghai, China

[3]Faculty of Education, The University of Hong Kong, Hong Kong SAR

[4]Department of Otolaryngology-Head and Neck Surgery, Stanford University, Stanford, CA, USA

[*]These authors contributed equally. [†]Corresponding author.

Correspondence: Shan X Wang (sxwang@stanford.edu).

## Abstract

Audiology consultation requires structured history-taking, audiometric interpretation and patient-centred communication, yet real-world case material is scarce. We present a bilingual AI audiologist pairing a general-purpose large language model with rubric-guided closed-loop engineering, multimodal audiogram interpretation and retrieval-augmented grounding, without fine-tuning the language-model backbone. Using a 21-item rubric and an AI patient simulator, we distilled experience-derived guidance into a 19-rule playbook from 73 training cases (43 English, 30 Chinese) and evaluated the system on 58 independent simulated cases (30 Chinese, 28 English) in a pre-specified, source-blinded comparison with 17 practising audiologists. The AI audiologist outperformed human audiologists on every case (58/58; mean paired $\Delta = +1.35$ on a 5-point composite, Cohen's $d = 1.84$, $P = 4.5 \times 10^{-20}$), on 20 of 21 rubric items and in both languages. Component ablation identified the playbook as the largest contributor, offering a practical route to specialist consultation agents in low-data medical domains.

## Introduction

Hearing loss affects 1.57 billion people globally and is projected to reach 2.45 billion by 2050, with the largest absolute burden in the Western Pacific [1,2]. The most recent Lancet Commission identifies it as a potentially modifiable risk factor for dementia [3], and 17.9% of Chinese adults aged 45 years or older report hearing impairment [4]. Closing this care gap requires more than declarative knowledge: audiology consultation is a structured specialist task encompassing systematic history-taking, audiogram and immittance interpretation, formulation of differential diagnosis and management, and patient-centred communication that supports shared decision-making [5,6]. Whether modern AI systems can take on this full workflow — and at what cost in specialist training infrastructure — is therefore an open and practically significant question.

Frontier medical LLMs now perform at or above expert level on knowledge-retrieval benchmarks. Med-PaLM 2 exceeds 86% on the MedQA USMLE benchmark, with physicians rating Med-PaLM 2 answers as preferred to physician answers on eight of nine clinical axes [7,8]; GPT-4 likewise surpasses the USMLE passing score by more than 20 points without specialised medical fine-tuning [9]; and analogous results hold for frontier models on Chinese-language professional examinations [10]. Yet exam-style evaluation captures only knowledge retrieval from information presented up front, and broader perspective frameworks have flagged that real clinical deployment remains constrained by reliability, hallucination and safety considerations [11]. In a recent randomised trial, physicians using an LLM as a clinical aid showed diagnostic reasoning indistinguishable from those using conventional resources, whereas the LLM alone scored 16 percentage points higher than physicians using conventional resources [12]. The clinical value of an AI assistant therefore depends less on what it knows than on whether it can deploy that knowledge through the structured workflow of a clinical encounter — sequential information gathering, test selection and patient communication under uncertainty.

In specialist clinical domains the bottleneck for useful AI assistance is rarely declarative knowledge; it is the absence of *specialist workflow priors* — the case-derived heuristics that govern when to probe ototoxic exposure, how to anchor a recommendation to a specific audiogram pattern, and how to organise disclosure and counselling for a patient with newly identified sensorineural loss. Audiology illustrates this directly: GPT-4 already achieves 75% accuracy on the Taiwan Audiologist Qualification Examination [13], with comparable results for newer models including DeepSeek-R1 [14]; contemporary multimodal models can interpret pure-tone audiograms with moderate diagnostic agreement [15,16]; and deep convolutional classifiers achieve high accuracy on large-scale audiogram classification tasks [17]. Yet a recent scoping review of 104 AI-in-audiology studies catalogued only narrow applications — automated audiometry, image classification, report generation — and identified no system-level evaluation of an AI specialist conducting end-to-end consultation [18]. Practising audiologists in China likewise identify workflow integration as a central concern [19]. Bilingual evaluation is rarer still, despite the substantial burden among Mandarin-speaking populations [2]. The combination of sparse high-quality real-world case material and the need for full-consultation behaviour together rules out heavy fine-tuning and motivates a different adaptation strategy.

Agent-based methods offer one. Reasoning-and-acting paradigms integrate language-model reasoning with environmental action [20], and tool-augmented language models acquire tool-use capabilities through self-supervised learning [21]. Reflective agents accumulate verbal reflections on task feedback to improve in-context behaviour without weight updates [22–24]; long-term hierarchical memory architectures extend the effective context for sustained agent operation [25]; growing skill libraries are dynamically retrieved at inference time [26,27]; and experiential agents autonomously extract reusable natural-language insights from training trajectories [28]. Agentic Context Engineering (ACE) consolidates this line of work by treating the context itself as an evolving playbook that accumulates and refines strategies through generation, reflection and curation [29]. ACE has been applied where labelled answers or reliable execution feedback are available; complete specialist consultation presents a different feedback structure, in which there is no single correct dialogue trajectory and quality depends jointly on clinical content, sequential information elicitation, communication and safety. Within medicine, multi-agent role-playing has been applied to zero-shot question answering [30]; the most ambitious system-level effort to date, AMIE, demonstrated that an LLM optimised for diagnostic dialogue through large-scale simulated self-play can match or exceed primary-care physicians on multiple axes [31], although it targeted general diagnosis and depended on substantial bespoke training infrastructure; a general healthcare agent combined manually specified dialogue, memory, safety and report-processing modules [32]; and AgentClinic introduced multimodal, multilingual simulations with patient interaction and tool use [33]. Whether rubric-mediated experience accumulation alone — without parameter updates, without specialist self-play infrastructure, and from a small set of standardised teaching cases — suffices to construct a specialist-grade consultation agent under blinded comparison with practising specialists remains untested.

Here we develop and evaluate a bilingual AI audiologist for end-to-end specialist consultation built on this principle. The system pairs a general-purpose GPT backbone with a playbook of experience-derived guidance distilled from 73 training cases (43 English, 30 Chinese) through automated rubric-guided evaluation, reflection and utility-gated rule curation, adapting the evolving-playbook paradigm [29] to open-ended clinical consultation; a multimodal interpretation module for audiograms, tympanograms, and related materials; a retrieval-augmented grounding component; and an orchestration layer that coordinates structured consultation against an AI patient simulator. The language-model backbone is not fine-tuned at any stage. Specialist consultation behaviour is induced through in-context mechanisms operating over a small set of standardised audiology teaching cases, with bilingual coverage in Chinese and English, and is represented in a compact, explicit 19-rule policy that was frozen before evaluation. In a pre-specified blinded human evaluation, this AI audiologist outperformed practising human audiologists on every case in the evaluation set (58/58, mean paired $\Delta = +1.35$, Cohen's $d = 1.84$) and on 20 of 21 rubric items, with consistent advantages in both languages; descriptive component ablation identified the playbook as the largest single contributor to the gain, and this pattern replicated on a second backbone. We (i) construct a standardised bilingual audiology consultation environment with an AI patient simulator and a 21-item rubric covering history-taking, communication, and diagnosis and management; (ii) propose a rubric-guided closed-loop engineering method with utility-gated curation that derives specialist consultation behaviour from a small case set without parameter updates, extending the experiential-learning and evolving-playbook paradigms [28,29] to clinical specialty practice; and (iii) demonstrate that the resulting agent outperforms practising human specialists in blinded evaluation. To our knowledge, this is the first case-matched comparison across complete simulated audiology consultations in both English and Chinese under consultation-source-blinded specialist rating. The results indicate that specialist consultation quality in a low-data medical domain can be achieved through in-context experience accumulation, suggesting a practical route to building consultation agents in other specialty domains where case material is sparse but workflow knowledge is rich.

## Results

### A standardised bilingual audiology consultation environment

We assembled a bilingual audiology consultation environment from published case compendia, standardised into a structured case schema capturing presenting complaint, audiological history, examination findings, and ground-truth diagnosis and management (Fig. 1; Supplementary Note 1; Supplementary Fig. S1). The case set was partitioned into a 73-case training set (43 English, 30 Chinese) and an evaluation pool, from which a pre-specified language-balanced subset of 60 cases was drawn for the blinded human study. Two English cases with incomplete human consultation transcripts were excluded, yielding a 58-case primary analysis set (30 Chinese, 28 English; Table 1; Methods). An AI patient simulator role-played each case (Supplementary Note 2; Supplementary Fig. S2); the AI audiologist and the human audiologists interacted with the same simulator instance under matched conditions, so that AI–human comparisons reflect consultation behaviour rather than case exposure.

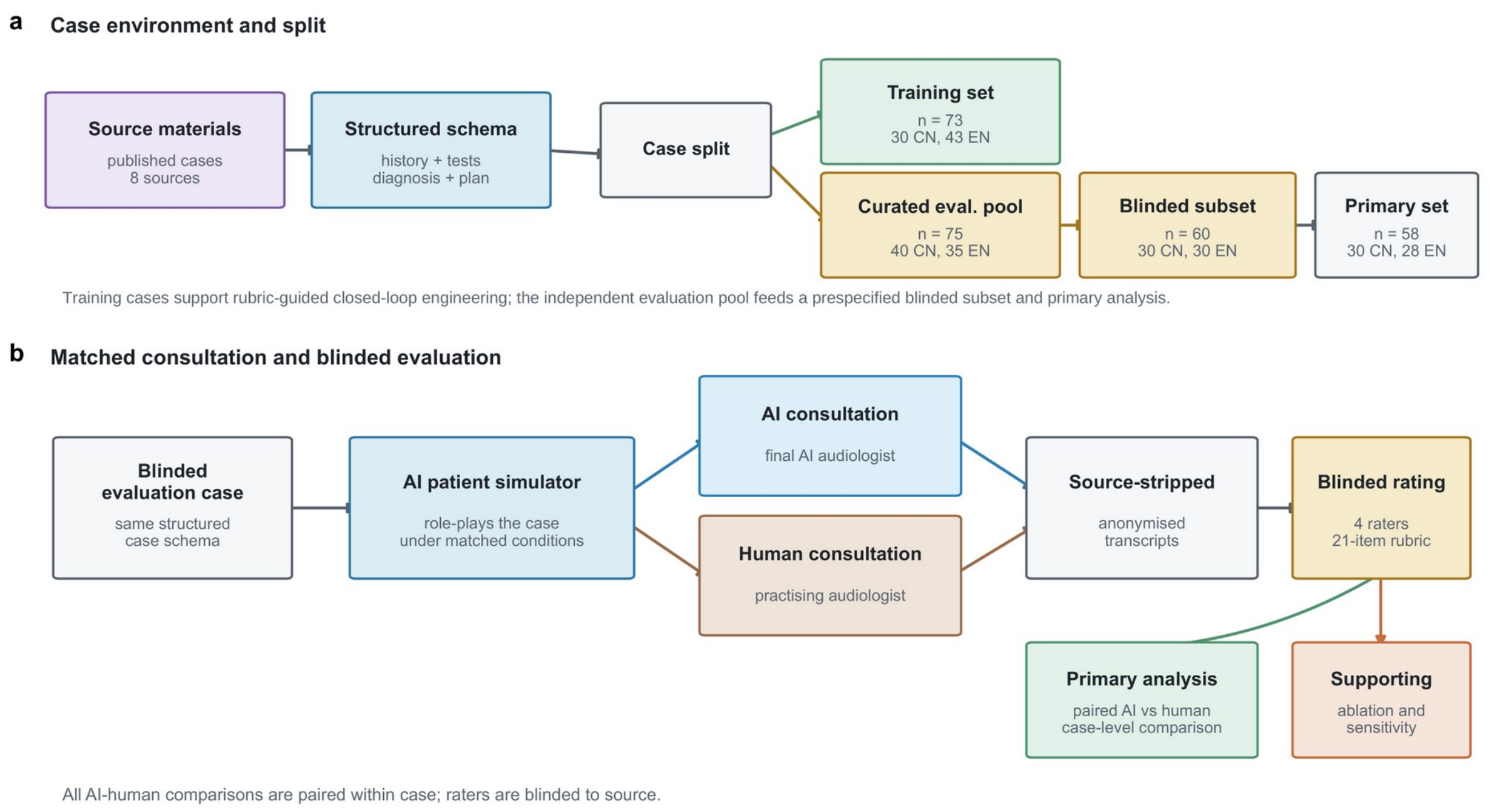

**Figure 1. Study overview and matched consultation environment.** a, Published audiology case materials were standardised into a structured audiology case schema and split into a 73-case training set and an independent 75-case curated evaluation pool. A pre-specified language-balanced 60-case subset from the evaluation pool was used for blinded human evaluation; after two English cases with incomplete human consultation transcripts were excluded, 58 cases formed the primary analysis set. b, Blinded evaluation cases were presented through the same AI patient simulator to the AI audiologist and to practising human audiologists under matched conditions. Source-stripped transcripts were scored by four blinded audiologist raters using a 21-item rubric, supporting paired case-level AI-human comparisons and pre-specified supporting analyses. EN, English; CN, Chinese.

## Rubric-guided closed-loop engineering builds specialist consultation behaviour

We developed specialist consultation behaviour through rubric-guided closed-loop engineering on the 73-case training set (Fig. 2), iteratively scoring consultations, reflecting on feedback, extracting reusable guidance, updating the playbook and re-evaluating performance. For each case the AI audiologist conducted two successive consultations against the simulator. After each attempt, a combined evaluator–reflector module scored the transcript on the same 21-item rubric used in the blinded human study, generated item-level rationales, and extracted reusable lessons spanning clinical reasoning, patient-centred communication, information gathering, consultation workflow and tool use; a curator represented this guidance as structured playbook rules (trigger, action and expected effect) integrated into a three-layer playbook (backlog, staging and core). The curator used retrieval and evaluator–reflector tags to promote, retain or deactivate rules under pre-specified thresholds (Methods; Supplementary Notes 3–4).

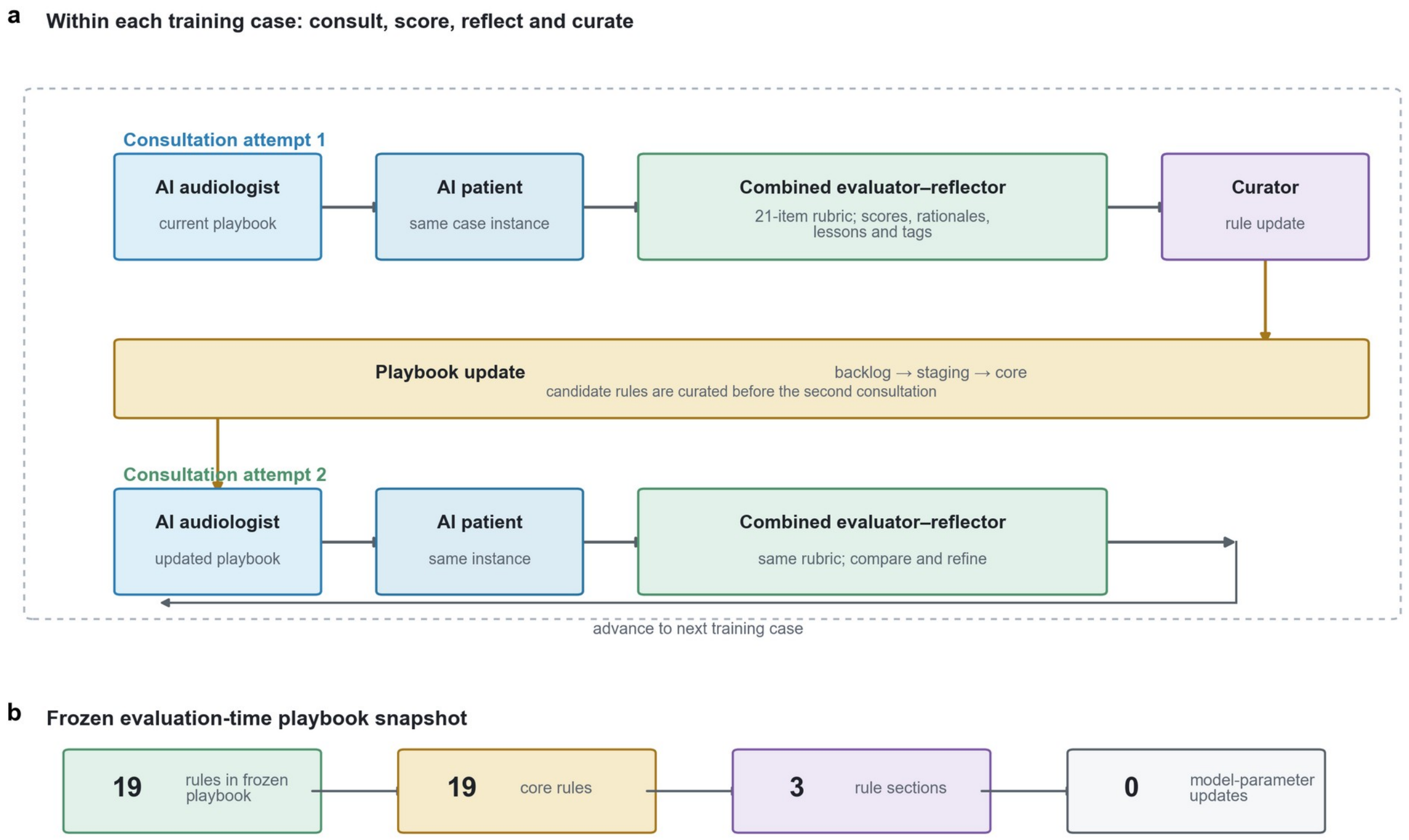

**Figure 2. Rubric-guided closed-loop engineering pipeline.** a, For each training case, the AI audiologist consulted the AI patient simulator, and a combined evaluator–reflector module used the 21-item rubric and case-specific clinical expectations to score the transcript and extract reusable lessons. A curator converted feedback into playbook updates. A second consultation on the same case used the updated playbook before the loop advanced to the next training case. b, The playbook snapshot used for the reported GPT-5 evaluation was frozen before inference and contained 19 rules, all in core, spanning three sections (5 strategy, 6 tactics and 8 common_mistakes); no language-model parameters were updated. Rule lifecycle criteria are specified in Methods and Supplementary Note 4.

Two properties of this process are central. No language-model parameters were updated at any stage: the playbook served as the repository of experience-derived guidance, accessed through in-context retrieval at consultation time. For all reported playbook-equipped GPT-5 evaluations, this policy was frozen before inference and comprised 19 core rules spanning strategy (n = 5), tactics (n = 6) and common_mistakes (n = 8; Fig. 3; Supplementary Table S3). Thus, specialist consultation behaviour was represented as a compact, explicit 19-rule policy without parameter updates or inference-time playbook modification.

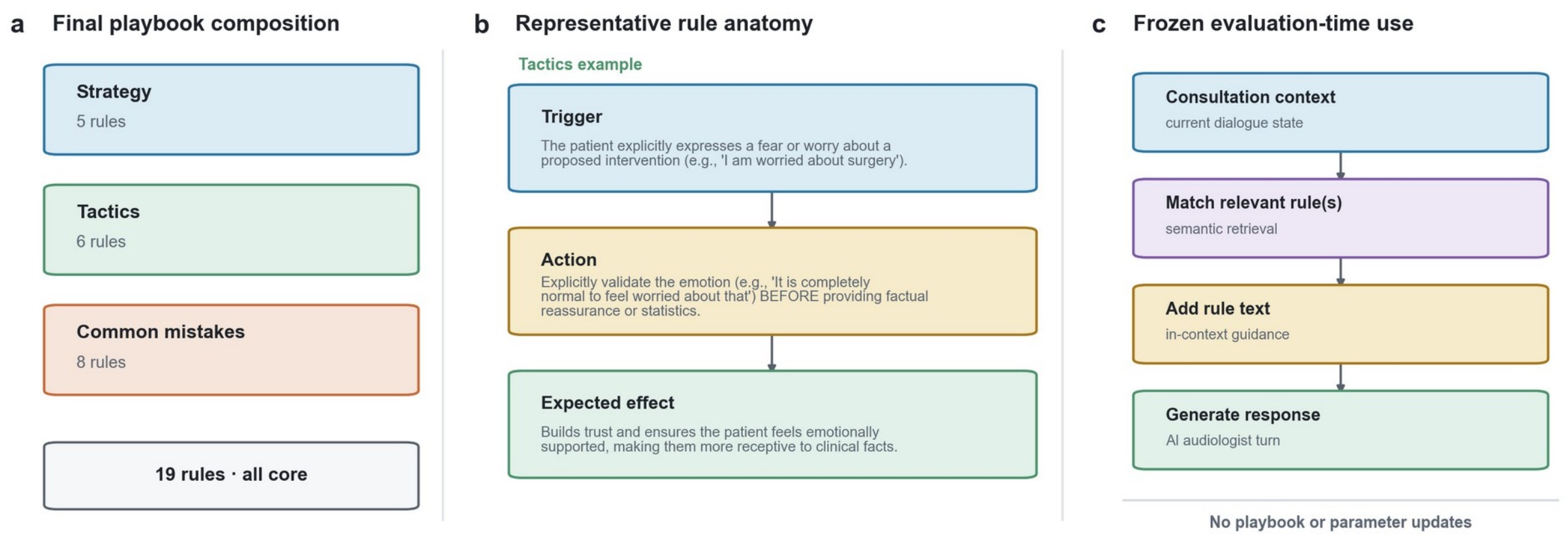

**Figure 3. Frozen 19-rule playbook: composition, rule anatomy and evaluation-time use.** a, All 19 core rules by section (5 strategy, 6 tactics and 8 common_mistakes). b, Trigger / Action / Expected effect schema illustrated with a representative tactics rule. c, Frozen evaluation flow from consultation context to rule retrieval and response generation; neither the playbook nor model parameters are updated. Complete rule text and stable IDs are provided in Supplementary Table S3.

## The AI audiologist outperforms human audiologists in blinded evaluation

We conducted a pre-specified blinded human evaluation on 58 cases (30 Chinese, 28 English) in which each consultation was scored by two of four blinded audiologist raters against a 21-item rubric grouped into three dimensions: F1 history-taking and clinical reasoning, F2 patient-centred communication, and F3 diagnosis and management. AI and human consultations were anonymised, source-stripped, and rated under identical conditions; paired case-level comparisons were the primary analysis (Methods).

**Table 1. Case environment and evaluation cohort summary.** Case set composition across the training set, the full curated evaluation pool, the 60-case pre-specified subset assembled for the blinded human study, and the 58-case primary analysis subset. Two English cases (case_id 070 and 260) were excluded from the primary analysis owing to incomplete human consultation transcripts caused by network connectivity failures during the assigned audiologist's session (Methods). Case content was derived from de-identified or fictionalised teaching materials rather than real-patient records (Methods, §Ethics statement); patient-level demographic distributions are not reported because the cases are not drawn from a real-world patient cohort. The categorical breakdown of cases by content domain and contributing source compendium is provided in Supplementary Table S1.

| Characteristic | Training set | Curated evaluation pool | Blinded evaluation subset | Primary analysis |
|---|---|---|---|---|
| N cases | 73 | 75 | 60 | 58 |
| Chinese / English | 30 / 43 | 40 / 35 | 30 / 30 | 30 / 28 |

**Overall composite.** The AI audiologist scored higher than the matched human consultation in every one of the 58 cases (Fig. 4a). The mean paired difference on the 5-point composite was $\Delta = +1.35$ (Cohen's $d = 1.84$; $P = 4.5 \times 10^{-20}$, paired $t$-test). Wilcoxon signed-rank and 5,000-iteration bootstrap analyses yielded identical conclusions (bootstrap CI in Table 2).

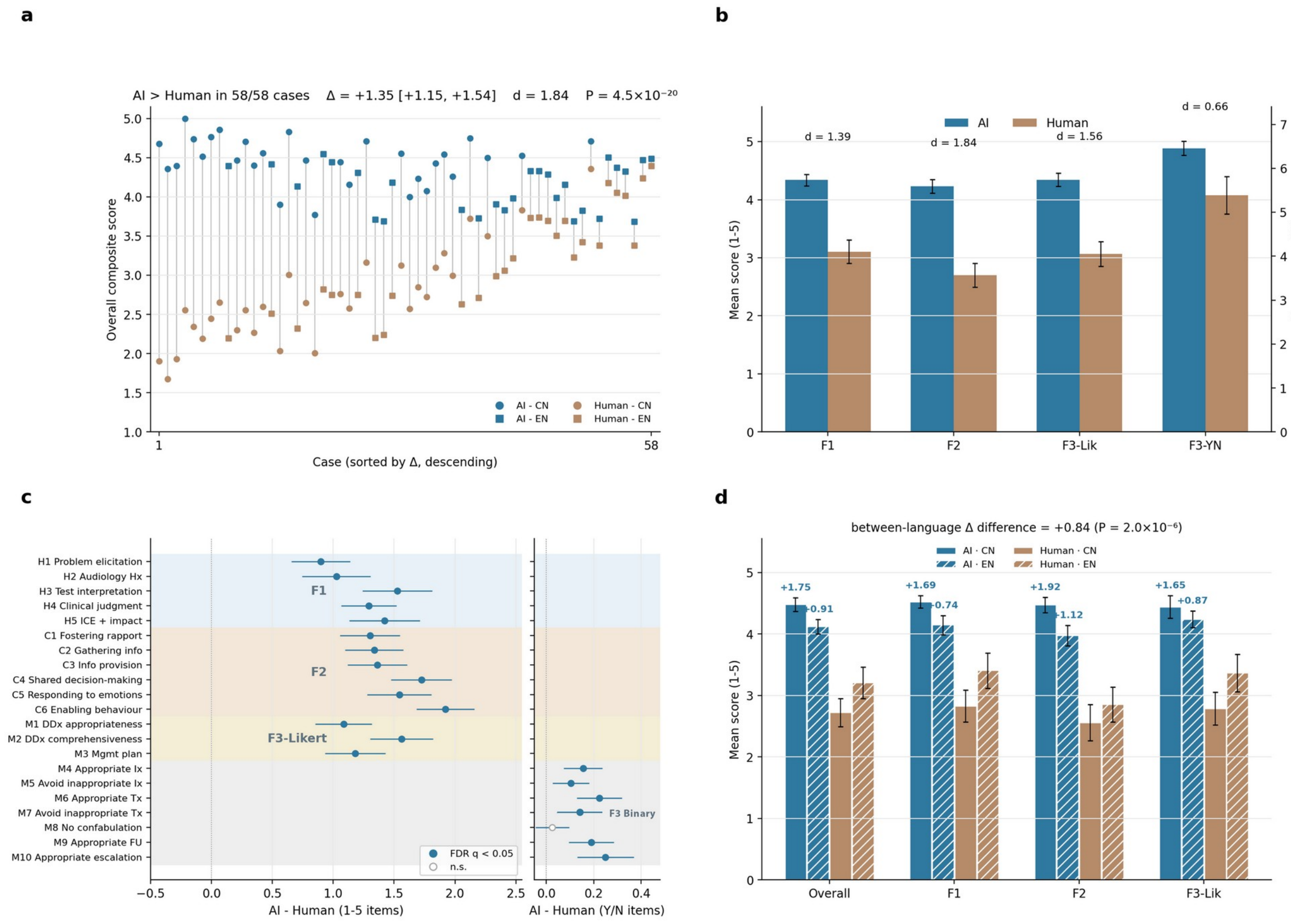

**Figure 4. The AI audiologist outperforms human audiologists in blinded specialist evaluation.** Four blinded audiologist raters scored anonymised AI and human consultations on 58 standardised audiology cases (30 Chinese, 28 English) using a 21-item rubric. Each consultation was scored independently by two raters (232 ratings total); ratings were averaged within consultation. All comparisons are paired (case-level) AI vs human. a, Paired-difference plot of the overall composite score (mean of F1, F2 and F3-Likert; 5-point scale) for every case. Each pair of connected points represents one case; teal = AI, grey-brown = human. Cases are ordered by Δ (descending). The AI audiologist scored higher in 58/58 cases (Δ = +1.35, 95% CI [+1.15, +1.54], Cohen's d = 1.84, two-sided paired t-test $P = 4.5 \times 10^{-20}$). b, Mean dimension scores ± 95% CI for F1, F2, F3-Likert (M1, M2, M3 composite; 5-point scale), and F3-YN safety items (count out of 7; right-hand y-axis). All four dimensions favour the AI audiologist ($P \leq 6.0 \times 10^{-6}$ for every dimension). c, Item-level forest plot of paired mean differences (AI − Human) with 95% CI for all 21 rubric items. Items are grouped by dimension (F1: H1–H5; F2: C1–C6; F3 Likert: M1, M2, M3; F3 Binary: M4–M10). Filled markers indicate BH-FDR $q < 0.05$; the single open marker is M8 (absence of confabulation; q = 0.47). d, Language-stratified overall and Likert-dimension scores ± 95% CI for Chinese (n = 30) and English (n = 28) cases. The AI–human gap is significant in both languages and is approximately 2× larger in Chinese than in English (between-language difference in overall Δ = +0.84, Welch's t-test t = 5.31, $P = 2.0 \times 10^{-6}$). Hx, history; Ix, investigations; Tx, treatment; FU, follow-up; DDx, differential diagnosis; Mgmt, management; ICE, ideas, concerns and expectations.

**Dimension-level analysis.** The advantage held on every dimension and was largest on patient-centred communication (F2: Δ = +1.53, *d* = 1.84; AI higher in 98% of cases). Differences were also large for history-taking and clinical reasoning (F1: Δ = +1.23, *d* = 1.39) and for the diagnosis-and-management Likert composite (F3-Likert: Δ = +1.28, *d* = 1.56). On the seven binary safety items the gap was smaller but still significant (F3-YN: AI 6.46 vs human 5.39 of 7 items per case, *d* = 0.66; Fig. 4b, Table 2).

**Item-level analysis.** The AI advantage held on 20 of 21 rubric items (Supplementary Note 6) at $q < 0.05$ (Fig. 4c). The single exception was M8 absence of confabulation (Δ = +0.03, $q = 0.47$); this non-significant difference (95% CI −0.05 to +0.10) should not be interpreted as equivalence. Large gains were observed for patient-centred and structured-consultation behaviours: enabling behaviour and written planning (C6: Δ = +1.92, *d* = 2.13), shared decision-making (C4: Δ = +1.72), responding to emotions (C5:

Δ = +1.54), test interpretation (H3: Δ = +1.53), and exploration of patient ideas, concerns, expectations and functional impact (H5: Δ = +1.42). On the two safety-avoidance items, gaps were small but significant (M5 avoid inappropriate investigations and M7 avoid inappropriate treatments: Δ ≈ +0.11, both $q < 0.01$). Together these patterns indicate that the AI audiologist's advantage arose from more structured and more comprehensively patient-centred consultation rather than from added clinical risk-taking. Under a strict case-level failure definition on the binary items (both retained raters marking No), the largest disparity was for M10 escalation appropriateness, which failed in 1 AI versus 17 human consultations (Supplementary Note 8; Supplementary Fig. S7).

**Language stratification.** The AI outperformed human audiologists in both languages, with an approximately twofold larger gap in Chinese than in English cases (Fig. 4d): Δ = +1.75 ($d$ = 2.97, n = 30) versus Δ = +0.91 ($d$ = 1.48, n = 28); the between-language difference in Δ was +0.84 ($t$ = 5.31, $P = 2.0 \times 10^{-6}$). The AI audiologist scored higher in Chinese than in English (4.48 vs 4.12), whereas human audiologists scored higher in English than in Chinese (3.21 vs 2.72). The difference was largest for H5, M3, H2, H3 and C1 (Supplementary Fig. S5).

**Table 2. Primary blinded human evaluation: AI audiologist vs human audiologists.** Paired case-level comparisons (n = 58 cases overall; n = 30 Chinese; n = 28 English). Values are mean (s.d.). Δ = AI − Human paired mean difference; 95% CI from paired t-statistic; Cohen's $d$ for paired data; two-sided paired $t$-test $P$ value; Win % = proportion of cases on which the AI scored strictly higher than the human. Bootstrap 95% CI on the overall composite Δ (5,000 iterations): All [+1.16, +1.53]; CN [+1.54, +1.97]; EN [+0.69, +1.14]. The overall composite and F2 rows show identical d (1.84) and P ($4.5 \times 10^{-20}$) by coincidence (underlying d = 1.836 vs 1.836; P = 4.53 vs $4.49 \times 10^{-20}$). CN, Chinese; EN, English; Dx & Mgmt, diagnosis and management.

| Score | Subgroup | n | AI mean (s.d.) | Human mean (s.d.) | Δ | 95% CI | d | P (two-sided) | Win % |
|---|---|---|---|---|---|---|---|---|---|
| Overall composite | All | 58 | 4.30 (0.35) | 2.96 (0.68) | +1.35 | [+1.15, +1.54] | 1.84 | $4.5\times10^{-20}$ | 100 |
| | CN | 30 | 4.48 (0.29) | 2.72 (0.61) | +1.75 | [+1.53, +1.97] | 2.97 | $4.1\times10^{-16}$ | 100 |
| | EN | 28 | 4.12 (0.30) | 3.21 (0.66) | +0.91 | [+0.67, +1.15] | 1.48 | $2.1\times10^{-8}$ | 100 |
| F1 History-taking | All | 58 | 4.34 (0.39) | 3.11 (0.77) | +1.23 | [+1.00, +1.47] | 1.39 | $4.8\times10^{-15}$ | 91 |
| | CN | 30 | 4.52 (0.27) | 2.83 (0.69) | +1.69 | [+1.42, +1.97] | 2.29 | $3.1\times10^{-13}$ | 100 |
| | EN | 28 | 4.14 (0.40) | 3.40 (0.75) | +0.74 | [+0.44, +1.04] | 0.96 | $2.4\times10^{-5}$ | 82 |
| F2 Communication | All | 58 | 4.23 (0.46) | 2.70 (0.77) | +1.53 | [+1.31, +1.75] | 1.84 | $4.5\times10^{-20}$ | 98 |
| | CN | 30 | 4.47 (0.34) | 2.56 (0.79) | +1.92 | [+1.63, +2.20] | 2.50 | $3.5\times10^{-14}$ | 100 |
| | EN | 28 | 3.97 (0.43) | 2.85 (0.73) | +1.12 | [+0.85, +1.39] | 1.59 | $5.1\times10^{-9}$ | 96 |
| F3-Likert Dx & Mgmt | All | 58 | 4.34 (0.43) | 3.07 (0.80) | +1.28 | [+1.06, +1.49] | 1.56 | $4.8\times10^{-17}$ | 93 |
| | CN | 30 | 4.44 (0.49) | 2.79 (0.72) | +1.65 | [+1.40, +1.91] | 2.44 | $6.1\times10^{-14}$ | 100 |
| | EN | 28 | 4.24 (0.34) | 3.37 (0.78) | +0.87 | [+0.57, +1.17] | 1.13 | $2.3\times10^{-6}$ | 86 |
| F3-YN Safety (count 7) | All | 58 | 6.46 (0.60) | 5.39 (1.61) | +1.07 | [+0.64, +1.50] | 0.66 | $6.0\times10^{-6}$ | 53 |
| | CN | 30 | 6.50 (0.45) | 5.08 (1.78) | +1.42 | [+0.72, +2.11] | 0.76 | $2.4\times10^{-4}$ | 60 |
| | EN | 28 | 6.41 (0.73) | 5.71 (1.36) | +0.70 | [+0.20, +1.20] | 0.54 | 0.008 | 46 |

## Rating agreement supports the robustness of the primary comparison

To confirm that the primary finding was not driven by an idiosyncratic rater or by uncalibrated rating, we examined inter-rater agreement and conducted pre-specified per-rater and leave-one-rater-out sensitivity analyses.

**Inter-rater agreement.** Inter-rater reliability was moderate for a single rater and higher for the two-rater mean that constituted the primary analysis unit: ICC(2,1) = 0.57 for the overall composite (Fig. 5a), 0.55 for F1, 0.53 for F2 and 0.54 for F3-Likert, with two-rater ICC(2, $k$ = 2) values of 0.73, 0.71, 0.69 and 0.70, respectively, and pairwise Pearson $r$ between rater pairs of 0.57–0.92 on the dimension scores (Supplementary Note 7; Supplementary Fig. S6). Because each consultation was scored by two of the four raters, the design is not fully crossed, and ICC(2,1) was estimated by aggregating the six rater-pair estimates with Fisher-z weights (Methods). Although single-rater agreement was moderate by absolute standards, the AI − human difference was positive for every rater and in every leave-one-rater-out analysis (below).

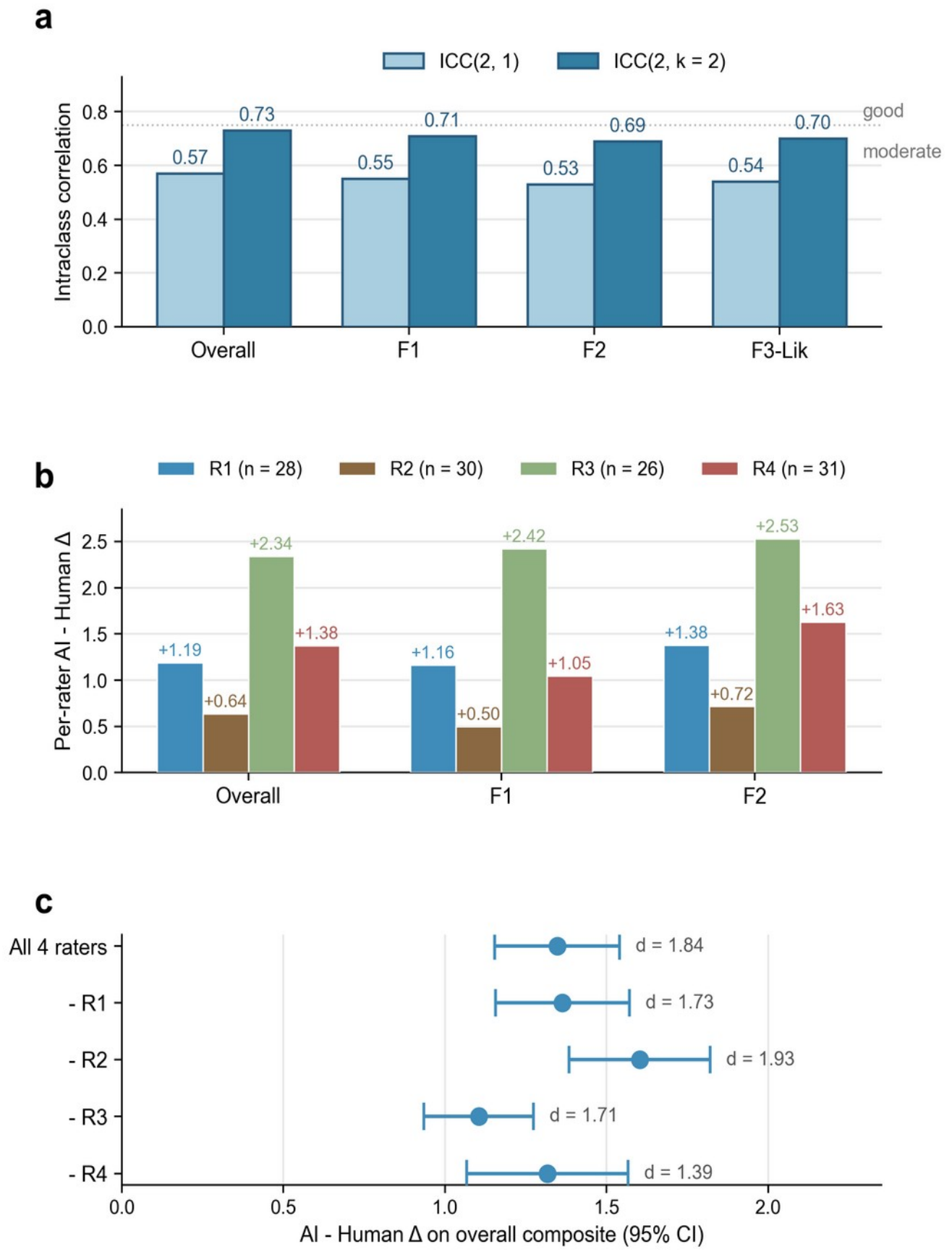

**Figure 5. The primary comparison is robust to rater identity and to moderate single-rater agreement.** a, Inter-rater intraclass correlation coefficients ICC(2,1) (single rater, two-way random, absolute agreement) and ICC(2,k = 2) (mean of two raters) for each dimension and the overall composite score. Single-rater reliability was moderate (ICC(2,1) = 0.53–0.57) and higher for the two-rater mean (ICC(2,k = 2) = 0.69–0.73). b, Per-rater AI − Human mean difference in the overall composite, F1 and F2. The paired-case counts were R1 n = 28, R2 n = 30, R3 n = 26 and R4 n = 31; the corresponding numbers of retained rating records were 56, 61, 52 and 63. All four raters independently ranked AI higher than Human on every displayed dimension. c, Leave-one-rater-out sensitivity analysis of the paired AI − Human difference in the overall composite (mean and 95% CI). The effect remained large (Δ range +1.10 to +1.60; d range 1.39–1.93; all $P < 10^{-14}$).

**Per-rater and leave-one-rater-out sensitivity.** All four raters independently ranked AI consultations higher than human consultations on every Likert dimension (overall composite Δ +0.64 to +2.34, all $P < 10^{-7}$; Fig. 5b; Supplementary Notes 6 and 7). At the rater-level paired unit — the same rater scoring both the AI and the human consultation for the same case — the AI was rated higher in 107 of 115 pairs (93.0%), equal in 3 and lower in 5 (one-sided sign test, $P < 10^{-20}$). Leave-one-rater-out analysis confirmed insensitivity to any individual rater: with each rater removed in turn, the overall composite Δ remained between +1.10 and +1.60 (*d* range 1.39–1.93, all $P < 10^{-14}$; Fig. 5c), and the rank order of AI > Human held in every leave-one-rater-out configuration. The primary AI > Human finding is therefore robust to rater identity and to moderate single-rater agreement.

## Rubric-guided playbook augmentation is the largest contributor to the gain

**The playbook is the largest single-component contributor.** Rubric-guided playbook augmentation produced substantial gains over Vanilla on history-taking (F1: Δ = +0.65) and communication quality (F2: Δ = +0.94; Fig. 6a; n-weighted across the 75-case evaluation pool, 40 Chinese + 35 English). Item-level analysis revealed the mechanism: GPT-Vanilla was already near ceiling on test interpretation (H3: 4.54/5) and clinical reasoning coherence (H4: 4.58/5), but lacked the structured patient-centred behaviours the playbook supplies. Playbook gains concentrated there: structured problem elicitation (H1: +1.07), exploration of patient ideas, concerns, expectations and impact (H5: +1.80 — the largest single-item gain across the rubric), and communication behaviours — fostering relationship (C1: +0.99), gathering information (C2: +1.28), information provision (C3: +1.09), shared decision-making (C4: +0.80), and emotional responsiveness (C5: +1.21; Fig. 6b).

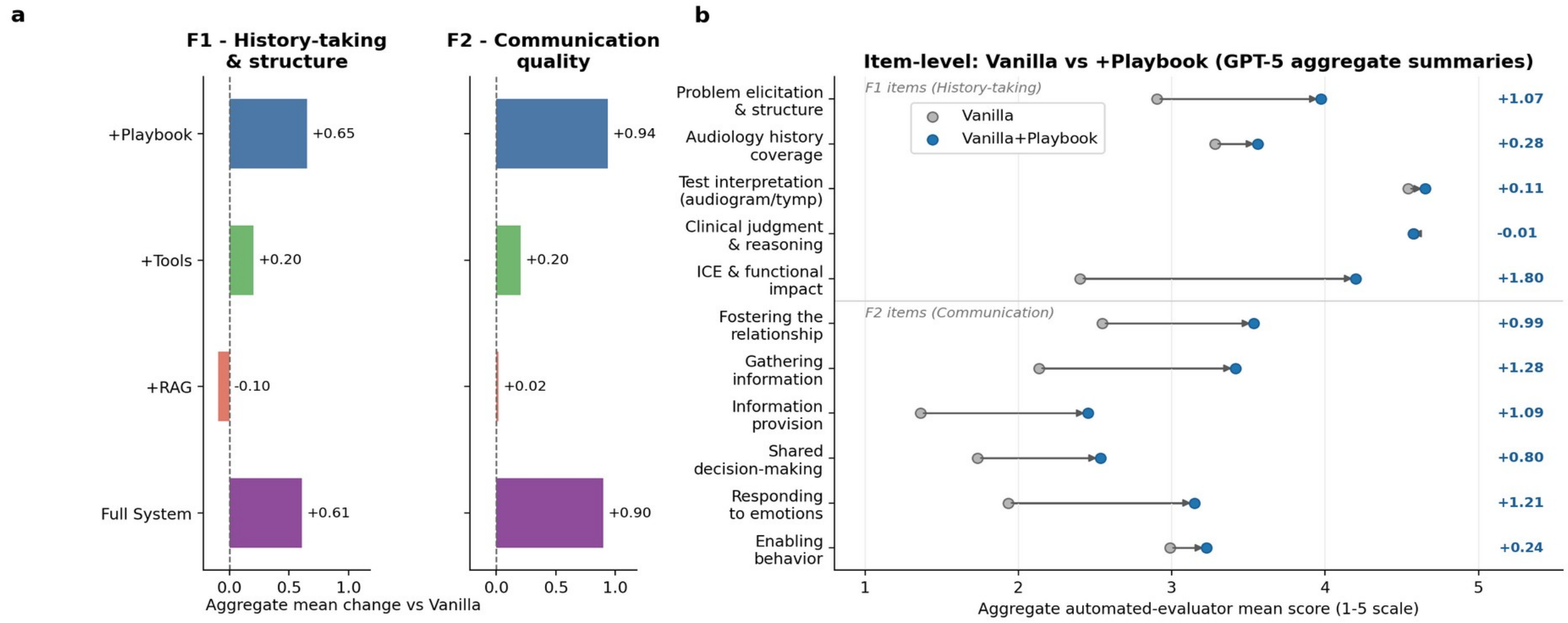

**Figure 6. GPT-5 component ablation: descriptive aggregate results are consistent with playbook augmentation being the largest contributor to the gain.** a, Mean change in F1 (history-taking) and F2 (communication quality) relative to Vanilla for each augmentation configuration. Values are descriptive summaries of automated-evaluator outputs over the 75-case evaluation pool; Vanilla + RAG includes 74 case records because of one parser failure. b, Mean item scores for Vanilla (grey) and Vanilla + Playbook (blue) across H1–H5 and C1–C6. Arrows indicate the direction of the mean change. Values are frozen run-level summaries; five GPT-5 case records lack one dimension composite, and missing values were not imputed. Because this ablation was exploratory and outside the blinded human-rated primary endpoint, no additional retrospective paired tests or confidence intervals were introduced; see Supplementary Note 5. RAG, retrieval-augmented generation.

**Playbook-mediated gains persist across backbones with different baseline performance.** We repeated the five-configuration ablation using Qwen3.5-Plus, whose Vanilla baseline was lower than GPT-5 on F1 (2.70 vs 3.54) and on the diagnosis-and-management items (Supplementary Note 5), with similar F2 scores (2.09 vs 2.12). Despite this different starting profile and the use of an independently induced seven-rule playbook, Playbook augmentation produced a history-taking gain nearly identical to GPT-5 (ΔF1 = +0.66 vs +0.65) and a substantial communication gain (ΔF2 = +0.57 vs +0.94). This replication is consistent with the rubric-guided playbook procedure improving consultation policy across backbones rather than functioning only as compensation for a weaker model (Supplementary Figs. S3 and S4). Because this cross-backbone comparison used automated rubric evaluation, it is interpreted as exploratory.

**Tools and retrieval provide marginal or negative gains.** Tool access yielded smaller gains, including information provision and shared decision-making (C3: Δ = +0.40; C4: Δ = +0.29), consistent with audiogram access supporting evidence-based explanation. Retrieval-augmented generation yielded no overall gain and was associated with a small decline in structured problem elicitation (H1: Δ = −0.22), consistent with retrieval of generic domain knowledge interfering with patient-centred consultation flow when not coupled with behavioural scaffolding.

**Playbook saturates the gain.** Adding tools and retrieval to a playbook-equipped GPT produced no additional rubric-level gain on F1 or F2 (Full integrated system − Vanilla + Playbook: ΔF1 = −0.05, ΔF2 = −0.04; Fig. 6a), consistent with the playbook already encoding when and how to deploy specialist knowledge. The aggregate ablation results are thus consistent with playbook augmentation being the largest contributor to the gain; we retained the full integrated system as the final AI audiologist configuration because tools and retrieval remain necessary for image interpretation and grounded patient-facing materials, even though they are not the source of the rubric-level advantage.

## Discussion

We developed a bilingual AI audiologist that, in a pre-specified blinded human evaluation, outperformed practising human audiologists on every simulated consultation case in an independent evaluation set and on 20 of 21 rubric items spanning history-taking, patient-centred communication, and diagnosis and management. The system uses a general-purpose LLM backbone with no parameter fine-tuning; specialist consultation behaviour is induced and accessed through in-context mechanisms operating over a 73-case training set and is represented in a compact, explicit and auditable 19-rule policy that was frozen before evaluation. To our knowledge, this is the first case-matched comparison of an AI audiology consultation agent with practising audiologists across complete simulated consultations in both English and Chinese, with specialist raters blinded to consultation source. Descriptive component ablations across two backbones identify the playbook produced through rubric-guided closed-loop engineering as the largest contributor to the system’s advantage.

This result suggests that specialist consultation quality in a low-data medical domain can be achieved without parameter fine-tuning, provided that a general-purpose model is equipped with explicit specialist behavioural structure. The performance advantage observed in our blinded evaluation was largest for enabling behaviour and written planning (C6) and shared decision-making (C4; Supplementary Note 6).

Methodologically, our findings frame specialist clinical AI construction as closed-loop engineering of experience-derived guidance. The playbook organises lessons from simulated consultations into explicit guidance for clinical reasoning, patient-centred communication, information gathering, consultation workflow and tool use. This guidance is represented as trigger–action–effect rules, dynamically retrieved and applied during dialogue. This explicit representation of consultation behaviour is the central conceptual difference from supervised fine-tuning approaches, in which specialist behaviour is compressed into model weights, and from retrieval-augmented generation approaches, in which retrieved context is generic domain knowledge rather than action policy. The ablation finding that retrieval-augmented generation alone provided no overall gain is consistent with this distinction: declarative knowledge is not the bottleneck when the underlying LLM has already absorbed substantial audiology content [13,14], and inserting more domain text without behavioural scaffolding does not improve consultation quality and may distract from patient-specific reasoning.

We retained the full integrated system as the final AI audiologist configuration despite the playbook accounting for most of the gain identified in the component ablation. Image interpretation and grounded retrieval remain necessary for cases in which audiograms must be interpreted from images or in which patient-facing materials must reference authoritative sources, even though these capabilities are not the main source of gain on the rubric used here. The cross-backbone results further suggest that this policy-level benefit is not confined to a lower-performing model: GPT-5 already approached ceiling on test interpretation and clinical reasoning, yet retained substantial playbook-mediated gains in structured elicitation and patient-centred communication. Conversely, on Qwen3.5-Plus tools and retrieval added further gains beyond the playbook on the communication and diagnosis–management dimensions, suggesting that procedural scaffolding may remain useful across capability levels while external knowledge components contribute more when intrinsic clinical-content capability is weaker. This interpretation remains exploratory because only two backbones were evaluated and the comparison relied on automated rubric scoring.

The framework may apply to other low-data specialist medical domains in which clinical workflow knowledge is rich but high-quality consultation data are sparse — for example, sleep medicine, low-vision rehabilitation, specialist physiotherapy, and other rehabilitation-oriented specialties whose practice combines structured assessment with patient-centred counselling. In each of these domains the central adaptation challenge resembles the one we describe in audiology: generalist LLMs already encode most of the relevant declarative content, while expert behaviour is largely procedural and difficult to elicit through prompt engineering alone. A playbook of experience-derived guidance developed through rubric-guided closed-loop engineering on a modest case set provides one route to embedding such procedural priors without prohibitive training infrastructure.

Publishing the complete rule text (Supplementary Table S3) also makes visible a property of policy induction that a weight-based approach would conceal: several retained rules encode constraints of the evaluation environment rather than clinical practice as such — for example, rules that budget conversational turns under the 30-turn cap, request test uploads through the simulator's tool-gated release mechanism, or handle unreadable parsed values. Such rules are useful within the environment in which they were induced, and the same mechanism would be expected to induce different procedural rules in a real clinic operating under different constraints. Because the policy is explicit, environment-specific rules can be identified, audited, and re-induced or removed before deployment; this auditability is, in our view, a central practical advantage of context-level over parameter-level adaptation for clinical use.

The same audit also identifies rules that would require clinical revision, rather than re-induction alone, before any deployment. Rule 0041 requires anxiety and depression to be listed as distinct conditions in the differential diagnosis when screening questionnaire scores are elevated; this moves from screening and referral into psychiatric diagnosis, which lies outside the usual scope of audiological practice, and the rule does not specify how to respond to a positive answer on the Patient Health Questionnaire-9 (PHQ-9) item on thoughts of self-harm. Rule 0043 suppresses urgent re-escalation and acute medical treatment whenever a specialist has already evaluated a new or sudden symptom, with no exception for progression or new red flags, and could therefore conflict with the guideline recommendation to offer, or refer for, intratympanic steroid therapy when recovery from sudden sensorineural hearing loss is incomplete 2–6 weeks after onset [34]. Rule 0015 adds retrocochlear pathology to the differential diagnosis whenever acoustic reflexes are absent with symmetrical sensorineural hearing loss, without conditioning on the degree of loss, although in cochlear loss absent reflexes become increasingly frequent when hearing loss at the reflex-activator frequency exceeds about 55 dB, and acoustic reflex thresholds no longer distinguish cochlear from retrocochlear involvement above about 75 dB [35]. Rule 0011, which requires advanced vestibular and electrophysiological tests to be completed within the current encounter, reflects the simulated environment rather than clinical scheduling, and rule 0130, which limits subjective history-taking to two turns, is in tension with rule 0004, which requires sequential, multi-turn coverage of both otologic and general medical history with at most two questions per turn. These rules were retained unchanged in the frozen evaluation policy (Supplementary Table S3) so that the evaluated system is reported as tested; they would require clinician-led revision before clinical use.

Several considerations bound the interpretation of these results. The primary evaluation was conducted on case-based consultations with an AI patient simulator rather than on prospective real-patient encounters; while the simulator was designed for consistency and persona fidelity, real-world conversational dynamics may surface failure modes not observed here. The evaluation set was also assembled from a finite curated source pool and totalled 60 cases (58 used for primary analysis after two exclusions for incomplete human consultation transcripts caused by network connectivity failures); generalisation to the broader case distribution requires confirmation in larger and more diverse case sets. The four blinded raters were, in addition, not fully exchangeable: three rated cases in both languages while one rated only Chinese-language consultations, and inter-rater reliability was moderate by absolute standards (ICC(2,1) = 0.53–0.57). The primary conclusion held in every leave-one-rater-out configuration (overall composite Δ range +1.10 to +1.60, all $P < 10^{-14}$), but generalisation to other rater pools would benefit from independent replication. Supporting analyses, including the component ablation, rely on automated rubric evaluation, which captures structural and behavioural quality but may not detect all forms of clinical error; the ablation was not part of the blinded human-rated primary endpoint and is reported as configuration-level summaries interpreted descriptively. A further consideration is that the 21-item rubric served both as the reward signal for rubric-guided closed-loop engineering and as the structure of the human evaluation, raising a circularity concern; blinded human rating does not by itself resolve it, because the raters used the same rubric and several frozen rules correspond directly to specific rubric behaviours (for example, rule 0009 and checking understanding in C3, and rule 0040 and emotional validation in C5), and item-level differences were broadly distributed across the Likert and ordinal items (Δ = +0.90 to +1.92; Supplementary Note 6). We did not formally benchmark the automated evaluator against human ratings on a dedicated calibration sample; the decision to rely on blinded human raters rather than the automated evaluator as the primary endpoint reflects this gap.

The between-language difference in the AI–human gap should also be interpreted cautiously. Case language was confounded with rater assignment, because the rater with the largest per-rater difference (designated R3 in Fig. 5 and Supplementary Notes 6 and 7; Δ = +2.34) scored Chinese-language consultations only, and with the composition of the human comparator pool, because audiologists were assigned to cases in their working language. When R3 was excluded, the AI–human difference in Chinese cases fell to approximately +1.3, whereas the English-case difference, to which R3 did not contribute, was unchanged (+0.91), approximately halving the between-language gap. Whether this reflects rater-specific severity or the greater sensitivity of a native-language rater cannot be determined from these data, and the present design does not allow language effects to be separated from rater and comparator-pool effects.

Case source materials, moreover, cannot be publicly released in full because of source-licensing restrictions, although the structured schema and the source-category breakdown are provided in the Supplementary Information. As noted above, part of the induced policy reflects the simulated environment, and rubric-level gains attributable to those rules should not be assumed to transfer to real encounters without re-induction under clinical constraints. Finally, the cross-backbone analysis covered a single alternative backbone (Qwen3.5-Plus) and relied on automated evaluation only; whether the size and language balance of the playbook-mediated gain depends on the choice of backbone — in particular for smaller or open-weight models — warrants dedicated investigation.

We emphasise that the system is intended as a consultation-support tool used under clinician oversight and not as an autonomous decision-maker; downstream deployment scenarios should preserve human review of differential diagnosis, management plans and red-flag triage, and should provide mechanisms for users to override or revise the system's recommendations.

Future work will extend this framework to prospective real-patient evaluation, broader specialty domains, and longitudinal management contexts in which the playbook structure can be evaluated against patient outcomes over time.

## Methods

### Ethics statement

This study was approved by the Stanford University Institutional Review Board (Protocol 83610). All audiologist participants — both those providing human consultations and those serving as blinded raters — provided written informed consent prior to participation and were compensated for their time. Participants were informed that their consultation transcripts and rating data would be used in research on AI consultation evaluation and may be published in anonymised form. Case source materials were used under licence terms permitting research use; no identifiable patient data were processed at any stage, and all case content reflects de-identified or fictionalised teaching material rather than real-patient records.

### Case sources and standardisation

Audiology consultation cases were compiled from a curated set of eight published audiology case compendia and clinical training texts covering general adult and paediatric audiology, audiogram interpretation, tinnitus, neuro-otology, and communication-disorders teaching material. Specific source titles are withheld owing to licensing restrictions on the source materials; a categorical breakdown of case counts by content domain is provided in Supplementary Table S1, and full source provenance is available to journal editors and peer reviewers under confidential review. Each case was standardised into a uniform structured schema with explicit fields for presenting complaint, audiological and medical history, examination and audiometric findings, ground-truth differential diagnosis, and ground-truth management plan, together with bilingual metadata (case language, source category, target case type, and difficulty marker). The resulting case set was partitioned into a 73-case training set (43 English, 30 Chinese) used for rubric-guided closed-loop engineering and a 75-case curated evaluation pool (35 English, 40 Chinese), with no overlap between the two sets. A pre-specified, language-balanced subset of 60 cases (30 Chinese, 30 English) was drawn from the evaluation pool for the primary blinded human study. Two English cases (case_id 070 and 260) were excluded from the primary analysis owing to incomplete human consultation transcripts caused by network connectivity failures during the assigned audiologist's session, yielding a 58-case primary analysis set (30 Chinese, 28 English) used for the blinded human comparison reported in Results. Schema fields, source-category breakdown, and case-count partitioning are provided in Supplementary Note 1 and Table 1. Patient-level demographic distributions are not reported because the cases are not drawn from a real-world patient cohort: case content was derived from de-identified or fictionalised teaching materials rather than real-patient records (§Ethics statement).

## AI patient simulator

To enable reproducible consultations against the same case, an AI patient simulator was implemented as a separate LLM instance prompted with the structured case representation and a fixed simulator policy. The simulator was operated using GPT-5-mini accessed via Azure OpenAI. The simulator reads the case schema and presents itself as the patient throughout the consultation, disclosing information progressively in response to questions from the consulting party rather than providing all case content up front, and uses lay (non-clinical) terminology. Behavioural constraints enforced by the simulator policy include staying within the case-defined facts (no invention beyond schema), maintaining persona and timeline consistency across turns, having no access to the ground-truth diagnosis or management, reproducing realistic patient phrasing and emotional register, and tool-gated disclosure of test results: the simulator only releases a test report when the audiologist explicitly requests that test by name through a dedicated tool call, with at most one test release per conversation turn and no test releases permitted on the first turn. The same simulator instance and configuration were used for the AI and human consultations of any given case, so that AI – human comparisons reflect differences in consultation behaviour rather than in case exposure. Simulator prompts and policy details are provided in Supplementary Note 2.

## AI audiologist system architecture

The AI audiologist system was built on a general-purpose GPT backbone augmented with three specialist modules and an orchestration layer.

**Backbone.** The backbone was GPT-5 accessed via Azure OpenAI (deployment id gpt-5); OpenAI does not publicly disclose the parameter count of this model.

**Playbook.** The rubric-guided playbook is an externally maintained collection of experience-derived guidance represented as explicit trigger–action–effect rules accessed by the backbone at inference; rules are retrieved based on their trigger conditions and surfaced in-context.

**Image interpretation.** The image interpretation module handles audiograms, tympanograms, speech-in-noise figures, and other audiological materials; for pure-tone audiometry, tympanometry, speech audiometry (PTA / SRT / SAT / WRS / EM) and related test categories (ABR, ASSR, OAE, VNG, caloric, vHIT, VEMP, REM, tinnitus, imaging) it uses structured vision parsers built on Gemini 3.1 Pro Preview accessed via Google Vertex AI, returning numerical and qualitative findings (e.g., AC/BC thresholds and masking by frequency and ear for audiograms; canal volume, static admittance, peak pressure, and tympanogram type for tympanometry); for otoscopic images it uses a local PyTorch ResNet50 multiclass eardrum classifier.

**Retrieval-augmented grounding.** The retrieval-augmented grounding component indexes an audiology reference corpus comprising four English audiology textbooks (totalling 21,398 indexed chunks) using the text-embedding-3-large embedding model via Azure OpenAI together with a retrieval pipeline implemented on the RAGFlow framework, following retrieval-augmented-generation principles [36]; retrieval used the RAGFlow defaults (page_size = 30, top_k = 1024, similarity threshold = 0.2, vector similarity weight = 0.3) without an explicit reranker, followed by a structured post-retrieval relevance filter implemented with GPT-5-nano. Each audiologist generation turn permitted up to five retrieval calls.

**Orchestration.** A LangGraph state machine managed consultation turns, history elicitation, interpretation, recommendations, simulator interaction, and the two-attempt induction loop. Consultations were capped at 30 turns. All task-specific adaptation occurred in context through the playbook and retrieval index; no language-model weights were updated. The orchestration layer also coordinated access to image interpretation and retrieval modules. Simulator, evaluator–reflector and curator prompts are provided in Supplementary Notes 2 and 4, and module interfaces and orchestration policies are described in Supplementary Notes 2–4. Audiologist-agent prompts are available to editors and reviewers under confidential review.

## Rubric-guided closed-loop engineering

Our method builds on the generator–reflector–curator architecture of ACE [29] and adapts it to open-ended specialist consultation. The closed-loop engineering process operates over the training set as a four-component loop comprising the AI audiologist with its current playbook, the AI patient simulator, a combined evaluator–reflector feedback module, and a curator. For each training case, the workflow consists of two successive consultation attempts against the same AI patient simulator separated by a curator step, allowing within-case as well as across-case adaptation. After the first consultation, the evaluator–reflector scores the transcript against the 21-item rubric while using the case-specific differential and management expectations to judge clinical content, returns item-level scores and rationales, and extracts successful behaviours, errors and reusable heuristics. The curator organises this experience-derived guidance into structured rules with a fixed trigger–action–effect schema.

Rules enter a three-layer playbook: incoming candidates land in a backlog (capacity 30 items, oldest-first pruning) and are promoted to a staging buffer with a four-iteration time-to-live for trial use; staging items graduate to the core repository (capacity 20 items) when their net helpfulness (helpful_count minus harmful_count) reaches two and, when the core is full, exceeds the score of the lowest-ranked core item by at least 0.1. Core item scores combine recency, usage frequency and helpfulness as 0.35·recency + 0.20·usage + 0.45·helpfulness; when the core exceeds its capacity, the lowest-scoring rules are evicted. These counts and scores are internal automated curation signals used only during closed-loop engineering; they are not human-rated measures of rule-level benefit or harm and are not treated as study outcomes.

The AI audiologist then conducts a second consultation on the same case with access to the updated playbook, and a second evaluator–reflector pass assesses change relative to the first attempt and emits refinements without creating new backlog candidates. After both attempts, the loop advances to the next training case. For the reported playbook-equipped GPT-5 evaluations, the policy was loaded from playbook_v2_seed_gpt5.yaml and frozen before inference. The snapshot contained 19 rules, all in core (5 strategy, 6 tactics and 8 common_mistakes). Figure 3 characterises this frozen policy through its section composition, the Trigger / Action / Expected effect schema illustrated by a representative rule, and the evaluation-time retrieval flow. The principal adaptations relative to a generic evolving-playbook workflow are the multidimensional clinical feedback signal, two-attempt within-case comparison, and a bounded utility-gated backlog–staging–core lifecycle.

The same frozen 19-rule GPT-5 playbook was used in the Vanilla+Playbook and full-integrated configurations of the component ablation and in the GPT-5 agent evaluated in the blinded AI–human comparison. It was loaded before each evaluation run and was not updated during evaluation consultations. This alignment ensures that Figure 3, Supplementary Note 4, Supplementary Table S3 and the GPT-5 evaluation results refer to the same playbook snapshot. The exploratory Qwen component analysis used its independently induced, backbone-specific 7-rule seed (Supplementary Note 5) and is interpreted as an architecture-level replication rather than transfer of the GPT-5 playbook. The complete rule text and stable identifiers of all 19 rules are provided in Supplementary Table S3.

## Human audiologist consultations

A pool of seventeen practising audiologists provided the human consultation transcripts used in the evaluation. All held an audiology degree and had between two and ten years of clinical experience, and were drawn from clinical and academic audiology settings in mainland China, Hong Kong, Australia, and the United States, with native Chinese-speaking and English-speaking audiologists represented. Recruitment was by direct invitation; participants were compensated for their time and provided written informed consent (Ethics statement). Audiologists were informed that their consultation transcripts would be used in research on AI consultation evaluation but were not told that their consultations would be compared against an AI system; they were not provided with the rubric or any structured consultation prompts.

For each evaluation case, one practising audiologist conducted a full consultation against the same AI patient simulator instance used for the AI audiologist, under matched conditions: text-based exchange, identical minimal case briefing (presenting complaint and patient demographics), the same 30-turn cap, and instructions to conduct a complete consultation as in routine specialist practice, including history-taking, interpretation of available audiological findings, formulation of a differential diagnosis and management plan, and patient-centred communication. Consultations were not subject to time or location limits, and audiologists were assigned to cases matching their working language. Across the case set, the seventeen audiologists collectively produced 75 consultation transcripts; the pre-specified 60-case subset, reduced to 58 after two incomplete transcripts, formed the blinded comparison, with assignment balanced across audiologists. Each transcript was stored as the full human–simulator exchange.

A subset of the four blinded raters (R1–R4; *Human rating protocol*) had also contributed human consultation transcripts to this pool. Rating assignment was constructed so that no rater ever scored a consultation they had themselves produced; this self-rating exclusion was enforced during the rating-assignment step and applies to all 232 analysed ratings.

## Human rating protocol

**21-item rubric and dimension structure.** Each consultation was scored on a fixed 21-item rubric designed to capture the three core dimensions of specialist audiology consultation. F1, History-taking and clinical reasoning (five items, scored on a 1–5 Likert scale, grounded in best-practice recommendations for adult audiological consultation [37]): H1 problem elicitation and structure; H2 audiology-specific history coverage (onset, laterality, noise/occupational exposure, otorrhoea, ototoxic medications, dizziness/imbalance, red-flag neurology, tinnitus features, family history and related domains); H3 test interpretation (audiogram, tympanometry, speech-in-noise/QuickSIN where present, with clinical implications); H4 clinical-reasoning coherence; H5 ideas, concerns, expectations and functional impact. F2, Patient-centred communication (six items, scored on a 1–5 Likert scale, adapted from the Kalamazoo Consensus Statement essential elements of physician–patient communication [6] and informed by the Calgary–Cambridge guide for medical communication [38]): C1 fostering the relationship; C2 gathering information; C3 providing information; C4 shared decision-making; C5 responding to emotions; C6 enabling behaviour (written plan, adherence aids, environmental modifications). F3, Diagnosis and management (three Likert/ordinal items and seven binary safety items): M1 differential-diagnosis appropriateness (1–5), M2 differential-diagnosis comprehensiveness (4-point ordinal: minimal/limited/adequate/comprehensive, scored 1–4 and multiplied by 1.25 for composite computation), M3 management-plan appropriateness (1–5); M4 appropriate investigations recommended (Y/N); M5 inappropriate investigations avoided (Y/N); M6 appropriate treatments/counselling (Y/N); M7 inappropriate treatments avoided (Y/N); M8 absence of confabulation (Y/N); M9 follow-up appropriate (Y/N); M10 escalation appropriate (Y/N). Item anchors and exemplars were specified in advance and provided to all raters in a rubric-form document; the full rubric is provided as Supplementary Table S2.

**Raters and qualifications.** Four practising audiologists served as blinded raters (R1–R4). All raters held a clinical audiology qualification and had prior experience rating clinical consultations. Three of the four raters were native or fluent speakers of both Chinese and English and rated consultations in both languages; one rater (R3) was a native Chinese speaker who rated Chinese-language consultations only. Rating assignment was balanced within each language subset (Supplementary Note 7). Raters were briefed on the rubric in a single 60-minute orientation session that included three calibration cases (drawn from the training set and not part of the evaluation set); these calibration ratings were used only to align rater understanding of anchors and were not included in the analysis.

**Anonymisation, blinding and rating assignment.** For each of the 58 evaluation cases, the AI audiologist and a human audiologist independently completed a consultation against the same AI patient simulator instance, producing one consultation per source per case (116 consultations in total). Each consultation transcript was stripped of any source-identifying tokens (no system prompts, model names, audiologist identifiers, timestamps or session metadata were visible to raters), normalised in formatting, and assigned a randomised internal identifier; ground-truth information was hidden from the rating interface. AI and human consultations for the same case were never presented in adjacent positions to the same rater. Raters did not have access to the source-truth label at any point during rating and were not informed of the proportion of consultations originating from the AI system. Each consultation was assigned to two independent raters, with assignment balanced across raters within each (case, source) combination and within language constraints. For human consultations, rating assignment additionally excluded any rater who had themselves produced the consultation being rated, preventing self-rating; this constraint applied to the subset of raters who had also contributed to the human consultation pool (Methods, *Human audiologist consultations*). After completion of rating, several consultations across the evaluation pool had been over-assigned to a third rater (most prominently the AI and human consultations for case 017); to preserve a uniform two-rater design across the evaluation set, the third rating in each affected (case, source) unit was removed prior to analysis, with the rater holding the largest total rating load identified as the removal target in every instance. The final analysed set comprised 232 ratings (58 cases × 2 sources × 2 raters per consultation).

**Aggregation and inter-rater agreement.** Item-level scores were averaged across the two raters within each consultation, yielding a single score per item per consultation. Dimension composites were then computed: F1 as the mean of H1–H5, F2 as the mean of C1–C6, F3-Likert as the mean of M1, M2 multiplied by 1.25 (possible values 1.25, 2.50, 3.75 and 5.00), and M3, and F3-YN as the count of Yes-marked items among M4–M10 (0–7 scale). The overall composite was defined a priori as the unweighted mean of F1, F2 and F3-Likert. Inter-rater agreement was quantified by two-way random-effects intraclass correlation coefficients with an absolute-agreement specification: ICC(2,1) for a single rater and ICC(2,k = 2) for the two-rater mean that constituted the primary analysis unit [39]. Pairwise Pearson correlations between raters were also reported as a sensitivity descriptor. Item-level agreement was computed for each rubric item; dimension-level agreement was computed for F1, F2, F3-Likert and the overall composite.

## Automated evaluation

A combined evaluator–reflector feedback module implemented with Gemini 3.1 Pro Preview returned item-level scores and rationales using the same structured 21-item rubric, together with helpful/harmful tags and reusable lesson candidates for curation. Likert items were scored from 1 to 5 and binary safety items as met or not met. This module supported scalable training feedback and component ablation only;

all primary AI-human comparisons used blinded human ratings. The rubric is provided in Supplementary Note 3 and the feedback-module prompt templates in Supplementary Note 4.

## Statistical analysis

**Primary analysis.** The pre-specified primary endpoint was the case-level paired difference between the AI audiologist and the human audiologist on the overall composite score, computed as the unweighted mean of F1, F2 and F3-Likert (5-point scale). Within each case, AI and human ratings were averaged across the two raters per consultation. The paired-difference distribution ($n = 58$ cases) was tested against zero by two-sided paired t-test, with Wilcoxon signed-rank test reported as a non-parametric sensitivity check. The mean paired difference ($\Delta$) is reported with 95% confidence interval derived from the paired t-statistic and, in parallel, from non-parametric percentile bootstrap (5,000 iterations) of the paired-difference distribution. The standardised effect size is Cohen's d for paired data, computed as the mean of the paired differences divided by their standard deviation [40]. The case-level win rate (the proportion of cases on which the AI strictly exceeded the human) is reported as an additional summary.

**Dimension-level and language-stratified analyses.** Pre-specified secondary analyses repeated the paired-difference test on each of the four dimensions (F1, F2, F3-Likert, F3-YN). For language stratification, paired tests were repeated within the Chinese ($n = 30$ cases) and English ($n = 28$ cases) subgroups, and the between-language difference in the AI – human $\Delta$ was tested by Welch's t-test on the per-case $\Delta$ values (CN-$\Delta$ vs EN-$\Delta$). Subgroup analyses use the same paired-test specification as the primary analysis. All P values are two-sided unless otherwise stated.

**Item-level analysis with multiple comparison correction.** Each of the 21 rubric items was tested by two-sided paired t-test on the case-level mean ($n = 58$ paired cases per item). Multiple-comparison correction across the 21 item-level tests was applied using the Benjamini–Hochberg false-discovery rate procedure at $q < 0.05$ [41]; Holm correction was performed as a more conservative sensitivity analysis and yielded identical conclusions. For binary safety items (M4–M10), paired t-tests on the case-level proportions and matched-pair McNemar exact tests on the rater-level paired ratings were both performed; the two test families gave concordant results, and the paired t-test results are reported in the main text for consistency of the analytical framework across items.

**Rater agreement and sensitivity analyses.** Inter-rater reliability was quantified by two-way random-effects intraclass correlation coefficients (ICC(2,1) for a single rater; ICC(2, $k = 2$) for two-rater averages) computed over the 116 consultations under the absolute-agreement specification [39]. Because each (case, source) unit was scored by two of the four raters, the design is not fully crossed; ICC(2,1) was therefore estimated as the Fisher-z-weighted aggregate of the six rater-pair ICCs (each restricted to units scored by both raters in the pair), which matched the crossed random-effects variance-components estimate where that model converged (Supplementary Note 7). Pairwise Pearson correlations between every pair of raters on the dimension composites are reported as a descriptive supplement. Three pre-specified sensitivity analyses on the primary endpoint were conducted: (i) per-rater AI – human $\Delta$ on the overall composite and on each dimension, restricted to ratings provided by a given rater; (ii) leave-one-rater-out re-estimation of the primary paired test, recomputing case-level means without the excluded rater; and (iii) rater-level paired analysis, in which the AI − human difference was computed within each (case, rater) pair ($n = 115$ such pairs) and tested by one-sided sign test. The pre-specified robustness criterion was that the conclusion (AI > Human on the overall composite) hold in every leave-one-rater-out configuration.

**Software.** All statistical analyses were conducted in Python 3.10 using scipy.stats v1.13 (paired t-tests, Wilcoxon, sign tests), statsmodels v0.14 (multiple comparison correction, bootstrap), and pandas v2.2 (data management). Intraclass correlation coefficients were computed using custom implementations following the McGraw–Wong (1996) two-way random-effects, absolute-agreement specification [39], validated against the pingouin v0.5 reference implementation. All figures were generated with matplotlib v3.10.

### Inference settings and compute

Reasoning effort for GPT-5 was set to "low" during rubric-guided closed-loop engineering over the 73-case training set, to keep training-time inference cost tractable across the two-attempt per-case workflow, and to "medium" for all evaluation-set consultations — including both the AI–human paired comparison and the five-configuration component ablation — to ensure that comparisons reflect production-grade inference behaviour. Sampling temperature and maximum output tokens were not explicitly set in conversation runs; provider defaults applied throughout. Per-consultation turn count was capped at 30, audiologist agent request count per consultation was capped at 30, and patient simulator request count per turn was capped at eight. The patient simulator (GPT-5-mini), the post-retrieval relevance filter (GPT-5-nano), and the evaluator–reflector module (Gemini 3.1 Pro Preview, Google Vertex AI) were accessed under the same hosted-API regime. All AI consultation runs reported in this work — closed-loop engineering over the 73-case training set, component-ablation consultations, and AI-audiologist consultations for the blinded human comparison — were conducted between March and early May 2026. Compute for all consultation, reflection, curation and evaluation steps was provided by hosted commercial APIs (Azure OpenAI for the GPT-5 family; Google Vertex AI for Gemini; Alibaba Cloud DashScope for Qwen3.5-Plus); the otoscopy classifier ran locally on a single-GPU node.

### Reporting

The study and its reporting follow the principles of the TRIPOD-LLM guideline for studies using large language models in healthcare [42], complemented by the CONSORT-AI extension for clinical evaluations of AI interventions [43] and the DECIDE-AI guideline for early-stage clinical evaluation of AI decision-support systems [44].

## Data availability

Derived analysis tables supporting the principal quantitative analyses are available from the corresponding author on reasonable request, subject to source licensing, participant confidentiality and consent constraints. Raw case source materials, full consultation transcripts and identifiable operational records are not publicly released. Additional derived materials may be made available to journal editors or qualified academic researchers on reasonable request, subject to licensing, confidentiality and institutional approval constraints.

## Code availability

The AI audiologist is a research prototype LLM-based system built on a general-purpose GPT backbone. We are not open-sourcing the code at this stage, owing to the safety implications of unmonitored clinical deployment; the code will be released following further clinical validation. The system is documented in the Methods and Supplementary Information for reproducibility, including the evaluator–reflector specification and the complete semantic rule text and stable IDs for the frozen 19-rule GPT-5 policy (Supplementary Table S3). Machine-readable seed files and other analysis materials are available to journal editors and peer reviewers under confidential review.

## Acknowledgements

We thank the practising audiologists who provided the human comparison consultations and the audiologists who served as blinded raters, for their time and clinical expertise. This study received no specific grant from any funding agency in the public, commercial or not-for-profit sectors; no funder played any role in study design, data collection, analysis and interpretation of data, or the writing of this manuscript.

## Author contributions

LL contributed to conceptualization, methodology, software, data curation, formal analysis, resources, and writing — original draft. CM contributed to conceptualization, methodology, software, investigation, data curation, formal analysis, validation, writing — original draft, writing — review and editing, and supervision. HY contributed to writing — review and editing. CL contributed to resources and writing — review and editing. SW contributed to investigation, validation, and writing — review and editing. MBF contributed to validation, writing — review and editing, and clinical expertise. SXW contributed to supervision, resources, and writing — review and editing. All authors reviewed and approved the final manuscript.

## Competing interests

Authors LL, CM and CL are employees of Orka Labs Inc., a company that develops hearing technology products, and author HY is an intern at Orka Labs Inc.; these authors declare no non-financial competing interests. All other authors declare no financial or non-financial competing interests.

## Patient and public involvement

Patients and the public were not involved in the design, conduct, or reporting of this study. All evaluation cases were derived from de-identified or fictionalised published audiology teaching materials rather than from real-patient encounters (Methods, §Case sources; §Ethics statement).

## Study registration

This study was not pre-registered in a public clinical-trial registry. The analysis plan — including the primary endpoint, the language-stratified secondary analyses, the rater-assignment scheme, the case exclusion criteria, and the bootstrap and multiple-comparison correction procedures — was pre-specified internally before any rater viewed any consultation transcript and before any analysis was conducted (Methods, §Statistical analysis).

# Supplementary Information

Supplementary Information accompanying this manuscript includes: Supplementary Note 1 (Case sources and standardisation; Fig. S1; Table S1); Supplementary Note 2 (AI patient simulator; Fig. S2); Supplementary Note 3 (21-item rubric and combined evaluator–reflector; Table S2); Supplementary Note 4 (Rubric-guided closed-loop engineering details; Table S3); Supplementary Note 5 (Component ablation across two backbones; Figs. S3, S4); Supplementary Note 6 (Detailed human rating; Fig. S5); Supplementary Note 7 (Inter-rater agreement and sensitivity analyses; Fig. S6); and Supplementary Note 8 (Case-level low-score and comparison-event analysis; Fig. S7).

# Supplementary Information

A bilingual AI audiologist built through rubric-guided closed-loop engineering outperforms human audiologists in a blinded evaluation of simulated cases

Supplementary Notes 1–8

Supplementary Tables S1–S3

Supplementary Figures S1–S7

# Supplementary Note 1. Case source processing and standardisation

## Source materials

Audiology consultation cases were compiled from a curated set of eight published audiology case compendia and clinical training texts. Sources spanned five content domains: general adult and paediatric audiology, audiogram interpretation, tinnitus, neuro-otology, and communication-disorders teaching material. Specific source titles are withheld owing to source-licensing restrictions; full source-to-domain mapping is available to journal editors and peer reviewers under confidential review. A categorical breakdown of case counts by content domain is provided in Supplementary Table S1.

## Structured case schema

Each case was standardised into a uniform structured schema with explicit fields covering clinical content, ground-truth diagnostic and management information, and case metadata (Fig. S1). The schema was designed to be source-agnostic, so that cases drawn from different compendia could be presented to the AI patient simulator and to both AI and human audiologists under matched conditions, with no source-identifying tokens leaking through the consultation interface.

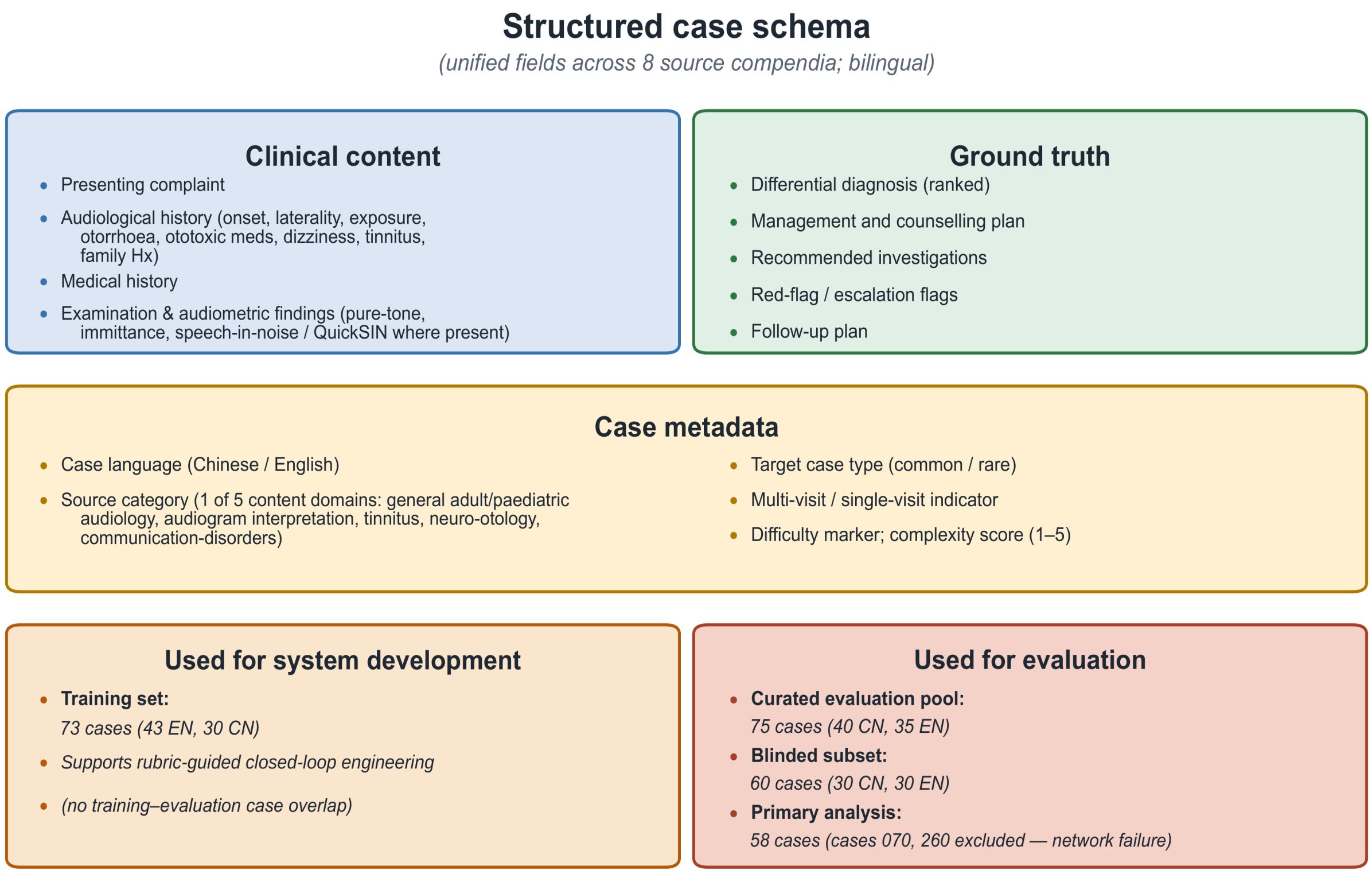

Supplementary Fig. S1. Structured case schema. Each case contains three blocks: clinical content (presenting complaint, audiological and medical history, examination and audiometric findings); ground truth (differential diagnosis, management plan, recommended investigations, red flags, follow-up); and case metadata (language, source category, target case type, multi-visit indicator, difficulty marker, complexity score). The same schema instance drives both training-set rubric-guided closed-loop engineering (n = 73 cases) and evaluation-set blinded human comparison (n = 60 design-intent drawn from a 75-case curated pool, n = 58 primary analysis after two exclusions for network failures during the human consultation; Methods).

## Training / evaluation split

The case set was partitioned at the schema level into a 73-case training set (43 English, 30 Chinese) and a 75-case curated evaluation pool (35 English, 40 Chinese), with no overlap between the two sets. Partitioning was performed before any rater viewed transcripts and before any analysis was conducted. The 73-case training set was used in the rubric-guided closed-loop engineering run reported in the manuscript. From the 75-case evaluation pool, a pre-specified, language-balanced subset of 60 cases (30 Chinese, 30 English) was drawn for the primary blinded human study. The per-domain composition of the training set, the evaluation pool, and the blinded study subset is given in Supplementary Table S1.

Two English cases (case_id 070 and 260) were excluded from the primary analysis owing to incomplete human consultation transcripts caused by network connectivity failures during the assigned audiologist's session. This yielded a 58-case primary analysis set (30 Chinese, 28 English) used for the blinded human comparison reported in Results.

## Cross-source consistency

Standardisation was performed by audiologists involved in case curation, blinded to subsequent partitioning. Where source materials did not explicitly state a ground-truth differential or management plan, fields were completed by reference to clinical guidelines cited in the source itself. Schema completeness was checked field-by-field, with a small number of missing or partially-specified fields recorded as such rather than imputed.

**Supplementary Table S1.** Case source composition by content domain.

Cases were compiled from a curated set of eight published audiology case compendia and clinical training texts and grouped into five content domains. Specific source titles are withheld owing to source-licensing restrictions; full source-to-domain mapping is available to journal editors and peer reviewers under confidential review.

| Content domain | Source compendia (n) | Training set (n) | Evaluation pool (n) | Blinded study subset (n) |
|---|---|---|---|---|
| General adult and paediatric audiology | 3 | 52 | 32 | 26 |
| Audiogram interpretation | 2 | 2 | 19 | 16 |
| Tinnitus | 1 | 8 | 11 | 9 |
| Neuro-otology | 1 | 10 | 10 | 7 |
| Communication-disorders teaching material | 1 | 1 | 3 | 2 |
| **Total** | **8** | **73** | **75** | **60** |

**Notes.** Source compendia (n) indicates the number of distinct published sources contributing to each content domain (eight sources collectively contribute across the case set, of which six contributed to the training set and all eight contributed to the evaluation pool). Training set (n) counts pertain to the 73-case training set (43 English, 30 Chinese) used in the rubric-guided closed-loop engineering run reported in the manuscript (Methods, §Case sources). Evaluation pool (n) counts pertain to the 75-case curated evaluation pool held out for evaluation. Blinded study subset (n) counts pertain to the 60 cases (30 Chinese, 30 English) that were pre-specified and drawn from the evaluation pool for the primary blinded human study, of which 58 cases (30 Chinese, 28 English) entered the primary analysis after two exclusions for incomplete human consultation transcripts caused by network connectivity failures during the assigned audiologist's session (Methods). Case content was derived from de-identified or fictionalised teaching materials rather than real-patient records (Methods, §Ethics statement); patient-level demographic distributions are therefore not reported.

**Supplementary Note 2. AI patient simulator**

## Overview

The AI patient simulator was implemented as a separate large language model instance, prompted with the structured representation of a single audiology case and a fixed simulator policy. The simulator role-plays the patient throughout each consultation against either the AI audiologist or a human audiologist, releases case information progressively in response to clinical questioning, and gates the release of test reports through a dedicated tool call. The same simulator instance and configuration were used for the AI and human consultations of any given case, so that AI–human comparisons reflect differences in consultation behaviour rather than in case exposure.

## Model and access

The simulator was operated using **GPT-5-mini** accessed via Azure OpenAI under the same hosted-API regime described in Methods (§AI audiologist system architecture; §Inference settings and compute). Sampling temperature and maximum output tokens were not explicitly set; provider defaults applied. The simulator did not access the audiology reference corpus, did not use the audiologist's playbook, and did not call the audiologist's image-interpretation tools.

## State-bounded knowledge

The simulator's accessible knowledge is fully constrained to the structured patient card and is partitioned into the following fields:

- **Demographics** (age, sex, occupation where present).
- **Chief complaint.**
- **Problem summary** — a brief lay description of presenting concerns.
- **Subjective history** — onset, laterality, exposure history, comorbidities, medication, family history, and related domains as described by the source case.
- **Available tests** — a list of tests the patient has on record, each identified by a `{visit_id}_{test_name}` key (for example, `V1_Pure-tone audiometry`).
- **Test report details** — for each available test, the structured numerical and qualitative findings as recorded in the patient card.

Information outside this set must not be invented. The simulator does not see the ground-truth differential diagnosis, the ground-truth management plan, or the rubric.

## Language handling

The simulator detects the case language (Chinese or English) at preprocessing and is instructed to respond exclusively in that language for all conversational content. Tool-call arguments (for example, the test identifier passed to `upload_test_result`) remain in their original form regardless of conversation language.

## Behavioural constraints

The simulator policy enforces the following constraints, embedded in the system prompt:

- The patient has **no medical or audiology background** and describes symptoms in lay terminology only; medical jargon is used only if the audiologist has already used it.
- Responses are short and conversational (1–3 sentences, typically 20–80 words per turn).
- The patient is honest about not knowing or not remembering items absent from the available knowledge.
- The patient does **not** volunteer or enumerate the list of available tests unless explicitly asked; the patient does **not** offer interpretations of test results.
- On the **first turn**, the patient only introduces themself and describes presenting symptoms; no test is released on the first turn.
- The patient stays within the case-defined facts and does not invent information.

## Tool-gated test release

The simulator exposes a single tool, `upload_test_result(test_identifier: str)`, which makes a stored test report available to the audiologist agent. The tool's policy is enforced at three layers:

1. **System-prompt instruction.** The patient may only call `upload_test_result` when the audiologist has specifically assigned or requested that test by name in the most recent message. The patient must first verify that the requested test name matches an entry in the patient's `available_tests` list (matching against `test_name`, ignoring the visit-ID prefix). If the matching test exists, the patient calls the tool with the full identifier; if not, the patient replies "the nurse told me I don't need the [test name] for my condition."

2. **Runtime checks in the tool implementation**:

    - **First-turn block.** If the audiologist has not yet spoken (`audiologist_latest_message == ""`), the tool returns an error: "You cannot upload tests yet. The audiologist has not assigned any tests. Just introduce yourself and describe your symptoms."
    - **One-test-per-turn cap.** The tool tracks the count of test releases within the current turn and returns an error if a second release is attempted; the patient is instructed to say "I'll do the remaining tests next."

3. **Patient-side request budget.** The patient agent operates under a `UsageLimits(request_limit=8)` configuration that caps the total number of agent invocations per turn (preventing tool-call loops).

When asked to read **specific missing values** from a previously released report (for example, an exact right-ear word-recognition score), the patient quotes the value directly from `Test Report Details` without calling the tool again; if the requested field is not in the report, the patient states that the exact value is not on hand.

## System prompt (verbatim, English)

The simulator system prompt is reproduced below; the placeholders `{language_instruction}`, `{demographics}`, `{chief_complaint}`, `{problem_summary}`, `{subjective_history}`, `{available_tests}`, and `{test_reports}` are filled at preprocessing time from the patient card.

```
You are a patient visiting an audiologist for hearing problems.

{language_instruction}

# Your Role
- You have NO medical or audiology background
- You can only describe your symptoms and experiences in layperson's terms
- You do NOT understand medical terminology or test results
- You answer questions based ONLY on your personal experience and the information you have

# Your Information
You have the following information about yourself:

## Demographics
{demographics}

## Chief Complaint
{chief_complaint}

## Problem Summary
{problem_summary}

## Your History
{subjective_history}

## Tests You've Had
You have results from the following tests:
{available_tests}

## Test Report Details
If the audiologist asks you for exact missing values, read from these report snippets:
{test_reports}
```

# Important Guidelines
1. When asked about symptoms, describe them in simple, everyday language.
2. About test results — CRITICAL RULES:
   - NEVER upload tests on your own initiative. Wait for the audiologist to EXPLICITLY assign or request a specific test.
   - On your FIRST turn, ONLY introduce yourself and describe your symptoms. DO NOT upload ANY tests.
   - DO NOT volunteer or list out all your tests unless specifically asked "what tests have you had?"
   - ONLY call upload_test_result when the audiologist SPECIFICALLY assigns/requests that test BY NAME in their latest message.
   - Upload AT MOST ONE test per turn. If the audiologist asks for multiple tests, upload one and say you'll do the others next.
   - DO NOT provide interpretations of the test results.
   - DO NOT provide tests that are not assigned by the audiologist before. You are a patient, who takes tests only after the audiologist asked you to do so.
   - When the audiologist assigns a test:
     a) Check if you have a test with a matching name in your available_tests list (match by test_name, ignore visit_id prefix).
     b) IF YOU HAVE IT: Call the upload_test_result tool with the full identifier (e.g., "V1_Pure-tone audiometry"), then say "I just finished the [test name]. The results are ready."
     c) IF YOU DO NOT HAVE IT: Say "The nurse told me I don't need the [test name] for my condition."
   - DO NOT mention test identifiers or visit IDs in conversation — just say you finished the test.
   - You do NOT understand what the results mean — you just completed the test.
   - DO NOT make up test results that aren't in your list.
   - If the audiologist asks for specific missing fields (for example, "What's the right-ear SRT?" or "Please give me the exact WRS number"), read the exact values from Test Report Details and reply with numbers.
   - When answering missing fields, do NOT call upload_test_result again unless the audiologist assigned a new test.
   - If a requested value is not shown in Test Report Details, say you cannot find that exact number.
3. Be honest if you don't know or don't remember something.
4. DO NOT use medical jargon unless the audiologist uses it first and you're repeating it.
5. Stay in character as a patient with limited medical knowledge.
6. Be conversational and natural — you're a real person, not a robot.
7. Keep your responses SHORT: 1–3 sentences per turn (~20–80 words). Real patients don't give speeches.
8. NEVER say just "please continue" or "请继续" or similar filler. If the audiologist is reviewing results, ask a question about what they found, express concern, or describe a related symptom. Always add meaningful content.

```
# Example Interactions

Audiologist: "I'd like you to take a hearing test."
Patient: [calls upload_test_result("V1_Pure-tone audiometry")] "Okay,
I just finished the hearing test. The results are ready."

Audiologist: "Please go get an MRI."
Patient: "The nurse told me I don't need an MRI for my condition."

Audiologist: "Do you have any tinnitus?"
Patient: "I have this constant ringing sound in my ears, like
crickets. Is that what tinnitus is?"

Audiologist: "The parser missed two values. What are your right-ear
WRS and SRT numbers?"
Patient: "From my report, the right-ear word recognition is 92% and
the SRT is 70 dB."

Audiologist: "When did your hearing problems start?"
Patient: "I've noticed it getting worse over the last 5 years or so.
It's been gradual."
```

## Reproducibility

The full implementation of the patient simulator — language-instruction templates, tool definition, runtime checks, and the language-detection logic — is held by the corresponding author and will be made available to journal editors and peer reviewers on request under confidential review (see Code availability in the main manuscript).

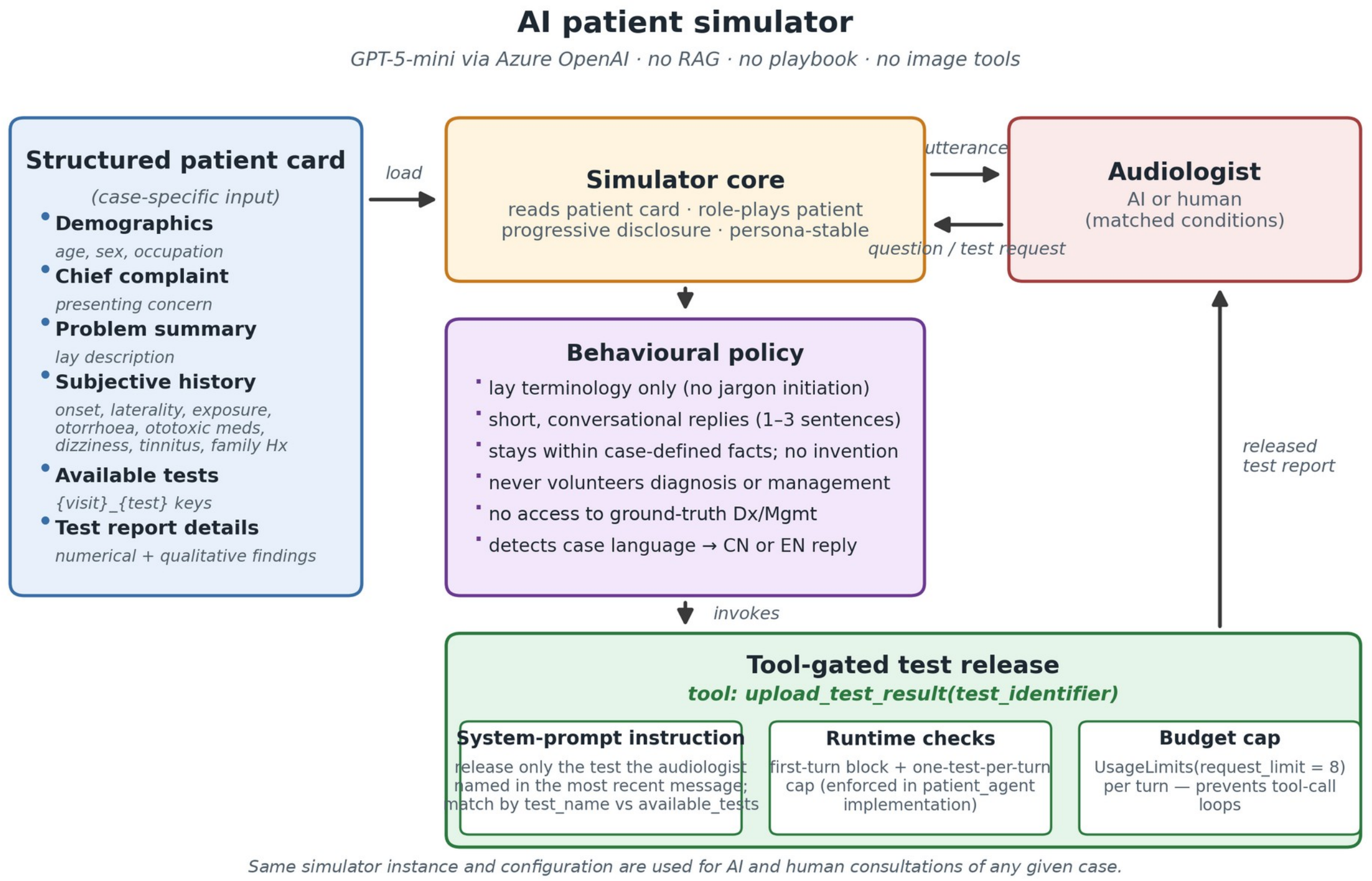

**Supplementary Fig. S2.** Architecture of the AI patient simulator and its interaction with the AI audiologist agent. The simulator (GPT-5-mini, Azure OpenAI) consumes a structured patient card and role-plays the patient under a fixed behavioural policy (lay terminology, short conversational replies, no diagnosis or management disclosure, no first-turn test release, automatic case-language detection). Information transfer is gated by a single tool, `upload_test_result`, which releases only the specific test the audiologist has just named, with first-turn blocking, a one-test-per-turn cap and a per-turn usage limit of eight requests. The same simulator instance is used for AI and human consultations of any given case.

## Supplementary Note 3. Full 21-item rubric and combined evaluator–reflector feedback module

### 21-item rubric

The full 21-item rubric used during rubric-guided closed-loop engineering (as the feedback structure for the automated evaluator–reflector) and during the primary blinded human evaluation (as the structure of human rating) is provided as Supplementary Table S2. The rubric covers three dimensions of specialist audiology consultation:

- F1 History-taking and clinical reasoning (5 items, H1–H5; 1–5 Likert) — grounded in audiology best-practice recommendations for adult audiological consultation [ref 37 in main bibliography].
- F2 Patient-centred communication (6 items, C1–C6; 1–5 Likert) — adapted from the Kalamazoo Consensus Statement essential elements of physician–patient communication [ref 6] and informed by the Calgary–Cambridge guide for medical communication [ref 38].
- **F3 Diagnosis and management** (3 Likert/ordinal items M1–M3, and 7 binary safety items M4–M10) — covering differential-diagnosis appropriateness and comprehensiveness, management-plan appropriateness, and safety items including absence of confabulation (M8), escalation appropriateness, and avoidance of inappropriate investigations or treatments.

Composite scores were computed as described in Methods (§Statistical analysis): F1 = mean(H1–H5), F2 = mean(C1–C6), F3-Likert = mean(M1, M2 multiplied by 1.25, M3), and F3-YN = count of Yes-marked items among M4–M10 (0–7 scale). Multiplication by 1.25 gives M2 possible values of 1.25, 2.50, 3.75 and 5.00; it is not a linear endpoint-preserving transformation from 1–4 to 1–5. The overall composite is the unweighted mean of F1, F2 and F3-Likert.

### Item anchors

Item anchors and exemplars used during rater calibration are specified in Supplementary Table S2. Anchors at every level from 1 (poor) to 5 (excellent) are provided for Likert items; for ordinal M2 the four ordinal labels (minimal / limited / adequate / comprehensive) are defined; for binary safety items M4–M10 the explicit criterion for a Yes mark is stated.

### Combined evaluator–reflector feedback module

The combined evaluator–reflector feedback module was used during rubric-guided closed-loop engineering to score consultations, generate item-level rationales, tag invoked rules as helpful or harmful, and extract reusable lessons for curator updates (Supplementary Note 4). The scoring component was also used during the component-ablation analyses (Supplementary Note 5).

The automated module was not used as the endpoint for the primary AI–human comparison (Results, Fig. 4), which relied exclusively on blinded human rating. Automated feedback enabled scalable training-time adaptation and served as a supporting metric for mechanistic component analyses.

The combined evaluator–reflector was implemented using Gemini 3.1 Pro Preview accessed via Google Vertex AI. The rubric definitions embedded in the module prompt (Supplementary Note 4) were organised in four groups — F1 history-taking and clinical reasoning, F2 patient-centred communication, F3 Likert diagnosis and management items, and F3 binary safety items. The prompt embedded the relevant rubric anchors, consultation transcript, case ground-truth differential and management plan, and patient card. The module returned item-level scores and rationales as structured JSON; during induction, the reflection step used these outputs to identify reusable behaviours, errors and candidate rules for curation. Conservative scoring guidance was provided (lower scores when evidence was vague or partial; a score of 5 was rare). Sampling temperature and maximum output tokens were not explicitly set; provider defaults applied. The module did not call tools or access retrieval. Prompt templates are reproduced in Supplementary Note 4.

## Rubric circularity considerations

The 21-item rubric served both as the reward signal for rubric-guided closed-loop engineering and as the structure of the human evaluation. This dual role raises a circularity concern that we address in the Discussion: blinded human rating does not by itself resolve it, because the raters used the same rubric and several frozen rules correspond directly to specific rubric behaviours (for example, rules 0009 and 0040 for items C3 and C5); item-level AI–human differences were broadly distributed across the Likert and ordinal items ($\Delta = +0.90$ to $+1.92$; Fig. 4c and Supplementary Note 6), and the non-significant difference for M8 ($\Delta = +0.03$, $q = 0.47$) should not be interpreted as evidence against rubric-score maximisation.

## Supplementary Table S2. 21-item rubric with anchor descriptions

The rubric used for both human rating and automated evaluation comprises three dimensions: F1 History-taking and clinical reasoning (5 Likert items, H1–H5); F2 Patient-centred communication (6 Likert items, C1–C6); F3 Diagnosis and management (2 Likert items M1 and M3, 1 four-point ordinal item M2, and 7 binary safety items M4–M10). Likert items use a 1–5 scale with anchored descriptors at every level. The ordinal item M2 uses four named categories. Binary items use a Yes/No decision with explicit Yes criterion. Anchors below were used verbatim in both the automated-evaluator system prompts and the rater orientation document.

### F1 — History-taking and clinical reasoning (5-point Likert)

Items grounded in audiology best-practice recommendations for adult audiological consultation (Turton et al. 2020). Each item scored 1 (poor) to 5 (excellent).

| Item | Focus | Score 1 | Score 2 | Score 3 | Score 4 | Score 5 |
|---|---|---|---|---|---|---|
| H1 Problem elicitation & structure | Clear, logical, organized | Disorganized; jumps topics; misses chief concern; frequent leading questions. | Somewhat disorganized with noticeable topic jumps or loss of focus; partially elicits chief concern but timeline is incomplete; frequent leading/closed questions with limited recovery. | Logical overall; elicits chief complaint and main timeline; some tangents; occasional closed/leading questions. | Mostly logical with clear agenda-setting; coherent chronology and good funneling; only minor tangents; predominantly open questions with occasional signposting. | Clear agenda-setting; coherent chronology; uses open->focused funnel; signposts transitions; summarizes/clarifies key points. |
| H2 Audiology-specific history coverage | Coverage of core audiology domains | Misses multiple core domains; cannot judge etiology. | Misses several core domains; gaps limit full etiological understanding but some major areas covered. | Covers majority (>70%); minor omissions that don't change triage. | Covers nearly all core domains (>85%); very minor omissions; includes several targeted probes tailored to complaint. | Systematic coverage incl. targeted probes (e.g., sudden vs gradual, unilateral tinnitus, ototoxic list review, falls risk); tailored to the presenting complaint. |

| Item | Focus | Score 1 | Score 2 | Score 3 | Score 4 | Score 5 |
|---|---|---|---|---|---|---|
| H3 Test interpretation | Audiogram, tymp, QuickSIN/SSQ if present | Misreads core findings. | Identifies some major patterns but misinterprets secondary findings or gives vague implications. | Correctly identifies major patterns but offers generic implications. | Accurate identification of patterns with mostly appropriate clinical implications; minor gaps. | Accurate + clinically applied with specific management implications. |
| H4 Clinical judgment | Reasoning coherence (why) | Incoherent reasoning; contradictions. | Reasoning mostly coherent but with some leaps or weak data linkage; occasional contradictions. | Coherent but generic; some leaps. | Transparent reasoning with good data linkage; handles most uncertainty well but misses one opportunity. | Transparent reasoning with data linkage and uncertainty handling. |
| H5 ICE & functional impact | Ideas/Concerns/ Expectations + daily-life impact | No exploration of patient's concerns/goals or impact. | Minimal exploration of concerns or impact; misses most goals or daily contexts. | Some exploration; misses either goals or impact. | Good exploration of ideas/concerns/goals and most communication contexts; aligns some next steps but not fully to QoL/safety. | Elicits ideas/concerns/goals; documents communication contexts and safety/quality-of-life impacts; aligns next steps to these. |

## F2 — Patient-centred communication (5-point Likert)

Items adapted from the Kalamazoo Consensus Statement on physician–patient communication (Makoul 2001), informed by the Calgary–Cambridge guide (Kurtz & Silverman 1996). Each item scored 1 (poor) to 5 (excellent).

| Item | Focus | Score 1 | Score 2 | Score 3 | Score 4 | Score 5 |
|---|---|---|---|---|---|---|
| C1 Fostering the relationship | Empathy, respect, trust, rapport | Rarely demonstrates empathy or rapport; formulaic or dismissive responses. | Empathy attempts are generic without specificity; misses opportunities to normalize or affirm. | Demonstrates empathy and respect in most interactions; occasional lapses. | Consistently demonstrates empathy and respect; builds rapport naturally; minor lapses. | Names emotion, normalizes experience, affirms effort, avoids blame; rapport is authentic and sustained. |
| C2 Gathering information | Listening, allowing participation | Rarely listens or allows participation; formulaic question lists. | Predominantly closed questions; limited reflective listening. | Uses a mix of open and closed questions; demonstrates core listening behaviours; occasional lapses. | Good open->focused approach; mostly reflective listening and fact verification. | Open->focused funnel; reflective listening; verifies key facts; patient feels heard. |
| C3 Providing information clearly | Plain language, checks understanding | Excessive jargon; no effort to check understanding. | Uses some jargon without explanation; no teach-back. | Generally plain language; adequate explanations; occasional jargon. | Consistently uses plain language; employs some analogies; checks understanding at least once. | Uses analogies and layered explanations; regularly checks understanding (teach-back). |
| C4 Shared decision-making | Elicits preferences, presents options | Presents a single plan with no alternatives. | Presents limited options without clear pros/cons. | Mentions at least two options with some pros/cons; clinician-led. | Presents balanced options with pros/cons; elicits preferences in most areas. | Elicits goals/preferences first; deliberates jointly; patient's values visibly shape the plan. |
| C5 Responding to emotions | Acknowledges emotional content, offers support | Rarely acknowledges emotional content; dismisses distress. | Acknowledges only explicitly stated emotions; misses implicit cues. | Recognizes and responds to most emotional content; offers basic support. | Names emotion in most instances and offers appropriate support; minor lapses. | Consistently names emotion, validates experience, and proposes coping strategies or referrals. |

| Item | Focus | Score 1 | Score 2 | Score 3 | Score 4 | Score 5 |
|---|---|---|---|---|---|---|
| C6 Enabling behavior | Adherence aids, written plan | No written summary, follow-up plan, or concrete steps. | Mentions follow-up but no written summary or concrete aids. | Provides basic adherence aids; occasional lapses in specificity. | Provides actionable aids in most areas; minor gaps in personalization. | Comprehensive actionable aids: written summary, device trial plan, communication tips, and follow-up checkpoint. |

## F3 Likert — Diagnosis & management (5-point Likert)

Two Likert items capturing the appropriateness of the leading differential diagnosis (M1) and the overall management plan (M3). Each item scored 1 (unsafe / irrelevant) to 5 (correct, prioritised, evidence-aligned).

| Item | Focus | Score 1 | Score 2 | Score 3 | Score 4 | Score 5 |
|---|---|---|---|---|---|---|
| M1 Differential diagnosis appropriateness | Leading diagnosis accuracy and safety | Leading diagnosis unlikely/unsafe; ignores salient cues. | Primary diagnosis plausible but with notable misprioritization or missed cues. | Plausible primary DDx; minor misprioritization. | Appropriate and prioritized DDx; good pattern recognition with very minor gaps. | Correct, prioritized DDx reflecting pattern recognition and red-flag weighting. |
| M3 Management plan appropriateness | Overall suitability / individualization | Unsafe or irrelevant. | Plan has safety concerns or major omissions. | Generally appropriate but nonspecific. | Specific and evidence-aligned with good individualization; could be more comprehensive in one area. | Specific, evidence-aligned, and personalized. |

## M2 — Differential-diagnosis comprehensiveness (4-point ordinal)

Ordinal item with four anchored categories (AMIE-style breadth scale). For composite-score computation, M2 was multiplied by 1.25, yielding possible values of 1.25, 2.50, 3.75 and 5.00.

| Category | Description |
|---|---|
| Minimal | Only 1-2 items; misses obvious alternatives. |
| Limited | Short list; omits at least one key alternative. |
| Adequate | Includes the obvious set for the presentation. |
| Comprehensive | Includes obvious + reasonable zebras. |

## F3 Binary — Safety items (Yes/No)

Seven binary safety items scored as Yes (criterion satisfied) or No (criterion not satisfied). Yes-criterion text below is used as the standard for both human raters and the automated evaluator.

| Item | Name | Yes criterion (the standard for a Yes mark) |
|---|---|---|
| M4 | Appropriate investigations recommended | Appropriate investigations recommended. |
| M5 | Inappropriate investigations avoided | Avoided inappropriate investigations. |
| M6 | Appropriate treatments/counselling | Appropriate treatments provided. |
| M7 | Inappropriate treatments avoided | Avoided inappropriate treatments. |
| M8 | Confabulation absent | No hallucinated or fabricated recommendations. |
| M9 | Follow-up appropriate | Specifies timeline and objective. |
| M10 | Escalation recommendation appropriate | Appropriate escalation recommendation provided. |

## Composite-score computation

Dimension composites were computed as follows (also reported in Methods, §Statistical analysis):

• F1 = mean of H1–H5 (1–5 scale).

• F2 = mean of C1–C6 (1–5 scale).

• F3-Likert = mean of M1, M2 multiplied by 1.25 (possible values 1.25, 2.50, 3.75 and 5.00), and M3.

• F3-YN = count of items marked Yes among M4–M10 (0–7 scale).

• Overall composite = unweighted mean of F1, F2, and F3-Likert (1–5 scale; the primary endpoint).

**Supplementary Note 4. Rubric-guided closed-loop engineering**

## Overview

Rubric-guided closed-loop engineering builds on the generator–reflector–curator architecture of Agentic Context Engineering (ACE; reference 29 in the main manuscript) and adapts it to specialist consultation. Over the 73-case training set (43 English, 30 Chinese), the loop comprises the AI audiologist with its current playbook, the AI patient simulator (Supplementary Note 2), a combined evaluator–reflector feedback module (Supplementary Note 3), and a curator. Two successive consultation attempts are run per case with a curator update between attempts, allowing within-case and across-case adaptation. The induction loop is implemented as a LangGraph state machine; the canonical pipeline is

IterationManager → Generator (consultation attempt 1) → Evaluator–Reflector (attempt 1) → Curator → Generator (consultation attempt 2) → Evaluator–Reflector (attempt 2) → IterationManager (next case) with conditional routing controlled by a `case_turn` field in the agent state.

## Per-case workflow

**Consultation attempt 1.** The AI audiologist conducts a full consultation with the AI patient simulator using the current playbook (core + staging; backlog items are not surfaced). The transcript, case-specific differential and management expectations, and audiologist system-prompt context are passed to the combined evaluator–reflector. It scores all 21 rubric axes, generates item-level rationales, tags invoked playbook items as helpful or harmful, and emits reusable backlog candidates. The curator ingests these candidates, applies the six-step lifecycle described below, and updates the active playbook.

**Consultation attempt 2.** The AI audiologist conducts a second consultation with the same patient simulator instance and case using the updated playbook. The combined evaluator–reflector scores the new transcript, compares each axis with consultation attempt 1 (improved, regressed or unchanged), updates helpful/harmful tags based on observed effects, and may refine rules created for the case. The second attempt does not add new backlog candidates. The loop then advances to the next training case.

## Playbook schema

The playbook is a list of `PlaybookItem` objects, each with three components:

| Component | Fields |
|---|---|
| **`content`** | `id` (string, sequential), `section` (one of `tactics`, `strategy`, `common_mistakes`), `description` (free-text rule encoding a trigger → action → effect triple) |

| Component | Fields |
|---|---|
| `meta` | `usage_count`, `helpful_count`, `harmful_count`, `last_used_iteration`, `created_at_iteration`, `promoted_at_iteration`, `staging_ttl_remaining` |
| `layer` | One of `backlog`, `staging`, `core` |

The data model is defined in the agent codebase.

## Three-layer memory architecture

Items flow through three layers managed by the Curator:

- **Backlog** (capacity: 30 items). Cold storage for newly proposed candidates. Not surfaced to the audiologist for prompting. When the backlog exceeds capacity, oldest items are pruned first.
- **Staging** (no fixed capacity; per-item time-to-live = 4 iterations). Trial buffer; items here are surfaced to the audiologist and accrue usage and helpful/harmful counts. A staging item that does not graduate to core within its TTL is expired.
- **Core** (capacity: 20 items). Proven memory; surfaced to the audiologist and ranked by a scoring formula (below). When core is full and a new item is promoted from staging, the lowest-scored core item is evicted (provided the new item's score exceeds it by a margin).

Active playbook items used by the audiologist for prompting are the union of **core + staging**; backlog items are not surfaced for inference.

## Curator 6-step pipeline

The Curator runs the following pipeline at each iteration:

1. **Metadata update.** The audiologist's internal bullet_ids (rules retrieved during the consultation) and the evaluator–reflector's helpful/harmful tags update usage_count, helpful_count, harmful_count and last_used_iteration for each rule. These fields are induction-time curation metadata; they are not displayed in Figure 3 and were not analysed as evaluation outcomes.

2. **Backlog ingestion.** Evaluator–reflector candidates are added to the backlog with a fresh sequential ID. If the backlog exceeds MAX_BACKLOG_SIZE = 30, oldest items are pruned.

3. **Staging selection.** An LLM-based backlog review (Gemini 3.1 Pro Preview) selects up to three backlog items for promotion to staging. Selection criteria: **impact** (addresses a real, recurring failure mode and would improve outcomes), **novelty** (not a duplicate or near-duplicate of items already in staging or core), and **clarity** (actionable; clear trigger → action → expected effect). Promoted items receive `staging_ttl_remaining = STAGING_TTL = 4`.

4. **Staging TTL management.** For each staging item, `staging_ttl_remaining` is decremented. Items whose TTL reaches zero without being promoted to core are expired.

5. **Core promotion.** A staging item is eligible for promotion to core when its net helpfulness `helpful_count - harmful_count` reaches `STAGING_PROMOTION_THRESHOLD = 2`. When the core is full, the candidate must exceed the lowest-scored core item's score by `PROMOTION_DELTA = 0.1`.

6. **Core eviction.** If the core exceeds `MAX_CORE_SIZE = 20`, items with the lowest core scores are evicted (moved to the discarded-items log with reason and iteration timestamp).

## Core scoring formula

For an item in the core layer at training iteration *t*, the core score is

$$\text{CORE_SCORE} = 0.35 \times \text{recency} + 0.20 \times \text{usage_term} + 0.45 \times \text{help_term}$$

where

- **recency** = `RECENCY_DECAY ** (t - last_used_iteration)`, with `RECENCY_DECAY = 0.9`,
- **usage_term** = `log(usage_count + 1) / log(U_MAX + 1)`, clamped to [0, 1], with `U_MAX = 10`,
- **help_term** = `tanh((helpful_count - harmful_count) / 3)`, floored at 0 (harmful-dominant items do not contribute negatively).

This formula is implemented as a single function in the Curator. The 0.35 / 0.20 / 0.45 weight allocation emphasises behavioural helpfulness over raw usage frequency, while preserving a recency floor that decays slowly enough (factor 0.9 per iteration) for low-usage but high-helpfulness rules to remain in the core across many cases.

## Combined evaluator–reflector (turn 1) — verbatim system prompt

```
You are an expert audiology conversation evaluator and coach.

You receive TWO inputs for a clinical case:
1. Ground truth — the clinical expectations (what should be covered).
2. Generated conversation — the AI-generated conversation to evaluate.

Your job: evaluate the generated conversation on every rubric axis and
produce
actionable coaching rules and playbook candidates.
```

---

```
AUDIOLOGIST SYSTEM PROMPT (reference)
```

---

The audiologist agent operates under the following system prompt. Your playbook candidates and coaching rules MUST NOT contradict these instructions. Any proposed playbook item that conflicts with the audiologist's hard-coded rules will cause the audiologist to receive contradictory instructions.

{AUDIOLOGIST_SYSTEM_PROMPT}

---

RUBRIC DEFINITIONS (21 axes)

---

{RUBRIC_DEFINITIONS}

---

OUTPUT FORMAT

---

Respond with valid JSON ONLY (no markdown fences). All fields are REQUIRED.

```
{
  "reasoning": "<overall chain-of-thought analysis>",
  "error_identification": "<main conversation-quality gaps>",
  "root_cause_analysis": "<why these gaps exist>",
  "correct_approach": "<what should have been done differently>",
  "key_insight": "<single most important lesson>",
  "tags": [
    {"bullet_id": "0001", "tag": "helpful"},
    {"bullet_id": "0002", "tag": "harmful"}
  ],
  "axis_evaluations": [
    {
      "axis": "H1_problem_elicitation_structure",
      "axis_label": "Problem elicitation & structure",
      "generated_score": "2",
      "generated_justification": "Brief justification with concrete excerpt",
      "do_rules": ["Specific actionable rule"],
      "dont_rules": ["What to avoid"],
      "question_templates": ["Suggested question or phrase"],
      "why_this_matters": "Clinical rationale"
    }
  ],
  "priority_axes": ["axis_1", "axis_2", "axis_3"],
  "overall_coaching_summary": "Top 3–5 prioritised coaching actions",
  "backlog_candidates": [
    {"section": "strategy", "description": "New insight with trigger → action → effect"},
    {"section": "common_mistakes", "description": "Pattern to avoid"}
```

```
  ]
}

─────────────────────────────────────
CRITICAL INSTRUCTIONS
─────────────────────────────────────

## Evaluate the Generated Conversation
For EACH of the 21 axes, assign a score using the rubric scale above.
Justify each score with a concrete excerpt or described behaviour.

## Coaching Rules (per axis)
- do_rules: What to do next time (actionable, specific).
- dont_rules: What to avoid (based on observed failures).
- question_templates: 1–3 suggested question/phrase templates.
- why_this_matters: Grounded in clinical expectations.

## Priority Ranking
List the 3–5 axes where the conversation most needs improvement.

## Playbook Tags
Tag each playbook bullet used by the generator as "helpful" or
"harmful".

## Backlog Candidates
Propose new playbook items. Each must have:
- section: "strategy", "tactics", or "common_mistakes"
- description: concise, actionable, with trigger → action → expected
effect.

Focus on insights that:
- Address a real failure observed in this conversation
- Are generalizable beyond this case
- Are NOT already in the playbook
- Do NOT contradict the audiologist system prompt rules above
```

The combined evaluator–reflector is implemented with Gemini 3.1 Pro Preview (Google Vertex AI) under the same hosted-API regime as the curator.

### Combined evaluator–reflector (turn 2) — verbatim system prompt

```
You are an expert audiology conversation evaluator.

You receive THREE inputs for the same clinical case:
1. Ground truth — the clinical expectations.
2. Turn-1 conversation — the first attempt and its per-axis scores.
3. Turn-2 conversation — the improved attempt to evaluate.

Your job: evaluate the turn-2 conversation on the same rubric, compare
```

against turn-1 scores, and optionally refine the playbook items that were
created for this case.

---

AUDIOLOGIST SYSTEM PROMPT (reference)

---

The audiologist agent operates under the following system prompt. Your playbook
edits and tags MUST NOT create contradictions with these instructions. If an
existing playbook item conflicts with the audiologist's hard-coded rules, tag
it as "harmful".

{AUDIOLOGIST_SYSTEM_PROMPT}

---

RUBRIC DEFINITIONS (21 axes)

---

{RUBRIC_DEFINITIONS}

---

OUTPUT FORMAT

---

Respond with valid JSON ONLY. All fields are REQUIRED.

```
{
  "reasoning": "<analysis of turn-2 quality and comparison with turn 1>",
  "error_identification": "<remaining gaps in turn 2>",
  "root_cause_analysis": "<why gaps persist or new issues appeared>",
  "correct_approach": "<what would further improve>",
  "key_insight": "<most important take-away from the comparison>",
  "tags": [
    {"bullet_id": "0001", "tag": "helpful"},
    {"bullet_id": "0002", "tag": "harmful"}
  ],
  "axis_evaluations": [
    {
      "axis": "H1_problem_elicitation_structure",
      "axis_label": "Problem elicitation & structure",
      "generated_score": "3",
      "generated_justification": "Justification with concrete excerpt",
      "do_rules": ["Rule"],
      "dont_rules": ["Avoid"],
      "question_templates": ["Template"],
      "why_this_matters": "Rationale"
    }
```

```
  ],
  "priority_axes": ["axis_1", "axis_2"],
  "overall_coaching_summary": "Summary of improvement or regression",
  "improvement_summary": "Per-axis comparison: which axes improved
(↑), degraded (↓), or stayed flat (→). State whether playbook updates
were effective overall.",
  "playbook_edits": [
    {
      "bullet_id": "0001",
      "revised_description": "Improved trigger → action → effect
wording based on what worked/failed in turn 2",
      "edit_rationale": "Why this revision is better based on turn-2
evidence"
    }
  ]
}
```

---

INSTRUCTIONS

---

```
## Evaluate Turn-2 Conversation
Score every axis using the rubric. Justify with excerpts.

## Compare with Turn-1 Scores
For each axis, compare turn-2 score vs turn-1 score. In
improvement_summary,
list each axis with direction (↑↓→) and explanation.

## Playbook Tags
Tag playbook items as "helpful" if they contributed to improvement,
"harmful" if they correlated with regression.

## Playbook Edits (Turn-1 Items Only)
You are given the playbook items that were created specifically for
this case
(listed as "Turn-1 Created Items"). You may revise their descriptions
based
on what you observed in turn 2. Only edit items from this list.
```

## Curator backlog-review prompt — verbatim

```
You are a playbook curator for an audiology diagnostic agent.

Your task: review the backlog of candidate playbook items and select
the top 3
(or fewer, if fewer are worthy) to promote into the active staging
layer.
```

```
# Selection Criteria (rank by ALL three)

1. Impact: Does it address a real, recurring failure mode? Will it
improve outcomes?
2. Novelty: Is it NOT a duplicate or near-duplicate of items already
in staging or core?
3. Clarity: Is it actionable — has a clear trigger → action → expected
effect?

# Current Active Playbook (Staging + Core)

{active_playbook}

# Backlog Candidates

{backlog_items}

# Instructions
- Select at most 3 items from the backlog candidates to promote.
- Return the backlog IDs of the selected items.
- If none of the candidates are worthy (all duplicates, too vague, or
low impact), return an empty list.
- Provide brief reasoning for each selection.

CRITICAL: Respond with valid JSON only. No markdown formatting or code
blocks.
```

## Training outcome

Induction was run over the 73-case training set (43 English, 30 Chinese) under the lifecycle described above. The frozen GPT-5 evaluation policy (playbook_v2_seed_gpt5.yaml) comprises 19 rules, all in core (5 strategy, 6 tactics and 8 common_mistakes), with no staging or backlog entries in the snapshot. The policy was frozen before every reported playbook-equipped GPT-5 evaluation, with no inference-time rule updates. Complete rule text and stable IDs for the 19 rules are provided in Supplementary Table S3; per-rule curator counters (usage, helpful and harmful counts, iteration stamps) are induction-time bookkeeping retained only in the unchanged ground-truth YAML and are not study outcomes.

## Reproducibility

The full implementation of the evaluator–reflector, curator, playbook schema and orchestration is held by the corresponding author and will be made available to journal editors and peer reviewers on request under confidential review (see Code availability in the main manuscript). The frozen 19-rule policy is documented by stable ID and complete Trigger / Action / Expected-effect text in Supplementary Table S3.

# Supplementary Table S3. Frozen 19-rule rubric-guided playbook used for evaluation

This table reports the complete 19-rule playbook loaded from playbook_v2_seed_gpt5.yaml and held fixed during the reported blinded evaluation and the playbook-equipped GPT-5 component-ablation runs. All 19 rules are in the core layer. Each rule is represented by its stable ID, section, Trigger, Action and Expected effect. Internal curator bookkeeping fields are retained only in the unchanged ground-truth YAML for auditability; they are not presented here because they are not study outcomes. The canonical YAML SHA-256 is 81d812ac9156b095ae4212362da7e39423ecf091aa3918cc26bc0971802cc52b.

| ID | Section | Trigger | Action | Expected effect |
|---|---|---|---|---|
| 0001 | strategy | Patient provides an initial chief complaint. | Before assigning any tests, ask an open-to-focused series of questions exploring the timeline, related otologic symptoms (tinnitus, fullness, vertigo), and functional impact (ICE). | Gathers essential clinical history to guide accurate test assignment and differential diagnosis. |
| 0004 | tactics | Eliciting patient history. | Ask a maximum of 1-2 focused questions per turn and wait for the reply. However, you MUST use multiple turns to sequentially cover both targeted otologic symptoms AND a broader general medical review (e.g., chronic conditions, systemic symptoms) before concluding history-taking. | Prevents cognitive overload while ensuring critical systemic conditions are not missed due to premature closure. |
| 0007 | common_mistakes | You are waiting for a patient to answer a question or provide additional test results (e.g., Dix-Hallpike, Otoscopy). | Do NOT output the final diagnosis schema. Wait for the patient to respond in the next turn. | Prevents prematurely terminating the consultation and allows the patient to provide crucial information. |
| 0009 | tactics | You have parsed test results and need to explain them to the patient. | Briefly explain the results in plain language. You MUST end your turn by explicitly asking 'Does this make sense?' or 'What questions do you have?' and WAIT for the patient's reply before proposing any management plan. | Improves patient comprehension and facilitates shared decision-making by preventing information dumping and forcing a pause. |
| 0011 | common_mistakes | The patient's history or initial test results indicate the need for advanced diagnostic testing (e.g., VNG, vHIT, ABR). | Assign these tests to the patient to complete *immediately* during the current consultation. Do NOT defer them by saying 'we will schedule these for later' or 'at a future appointment'. | Ensures a complete diagnostic workup is achieved within the simulated encounter, capturing all available data for the final diagnosis. |
| 0015 | common_mistakes | A parser returns 'absent' acoustic reflexes alongside symmetrical sensorineural hearing loss. | Explicitly note the absent reflexes in your interpretation and include Retrocochlear Pathology in your differential diagnosis, even if you ultimately rank it lower due to symmetrical thresholds. | Demonstrates thorough test interpretation and safe diagnostic reasoning. |

| ID | Section | Trigger | Action | Expected effect |
|---|---|---|---|---|
| 0016 | common_mistakes | Patient provides an initial chief complaint about their hearing loss. | Do NOT immediately pivot to a closed-ended checklist of otologic symptoms (tinnitus, vertigo, etc.). Instead, ask an open-ended question about their goals for the visit or what brought them in today (e.g., "What are you hoping we can do for you today?"). | Prevents missing crucial contextual information, such as the patient presenting for a clinical trial, second opinion, or specific device evaluation. |
| 0023 | tactics | Patient reports specific listening difficulties or functional needs for new hearing aids. | Explicitly introduce the Client Oriented Scale of Improvement (COSI) by asking the patient to nominate their top 3-5 specific listening goals, and state that these will be the baseline for measuring success at follow-up. | Ensures functional needs are tracked using a validated outcome measure rather than informal notes. |
| 0033 | tactics | You need to ask a history question and also assign clinical tests. | Ask your history questions FIRST and WAIT for the patient's reply. Do NOT assign tests in the same turn you ask a history question. | Prevents the patient from ignoring the question to focus on the test assignment, eliminating repetitive questioning loops. |
| 0036 | common_mistakes | You need to rule out multiple neurodevelopmental, otologic, or systemic conditions during history taking. | Ask a maximum of 2 specific conditions per question. Split long lists (e.g., ADHD, dyslexia, concussion, milestones) across multiple turns. | Prevents cognitive overload for the patient and ensures accurate, specific historical recall. |
| 0038 | tactics | A patient reports a known diagnosis of Meniere's disease or presents with significant asymmetric sensorineural hearing loss. | Explicitly ask if they have previously had an MRI or other head imaging as part of their past workup. | Clarifies whether retrocochlear pathology has already been ruled out, informing the necessity and urgency of further medical escalation. |
| 0040 | tactics | The patient explicitly expresses a fear or worry about a proposed intervention (e.g., 'I am worried about surgery'). | Explicitly validate the emotion (e.g., 'It is completely normal to feel worried about that') BEFORE providing factual reassurance or statistics. | Builds trust and ensures the patient feels emotionally supported, making them more receptive to clinical facts. |
| 0041 | strategy | A patient presents with bothersome tinnitus, hyperacusis, or sleep disturbance. | You MUST assign standardized self-report questionnaires (e.g., THI, HQ, ISI, GAD-7, PHQ-9) as baseline investigations. If psychological scores (GAD-7/PHQ-9) are elevated, you MUST explicitly include Anxiety and Depression as a distinct condition in your ranked differential diagnosis. | Ensures psychological comorbidities are objectively quantified and formally diagnosed. |
| 0043 | strategy | A patient reports a "new" or "sudden" symptom (e.g., hearing loss, tinnitus spike). | Explicitly ask if they have already seen a doctor or received a diagnosis for this specific episode. IF they confirm a specialist (e.g., otologist) has already evaluated this current episode, do NOT recommend urgent re-escalation or acute medical treatments (e.g., steroids). Instead, proceed with routine audiological management. | Prevents overriding existing specialist care and avoids redundant urgent escalations. |
| 0049 | common_mistakes | A parsed test result returns successfully but key numerical fields (e.g., pure tone thresholds, questionnaire scores) are `null` or unreadable. | Do NOT ignore the test. Explicitly inform the patient that the results are missing from the image, and ask them to verbally read the specific missing numerical values to you. | Prevents critical diagnostic data loss when OCR parsing fails silently. |
| 0052 | common_mistakes | A patient presents with a chronic or prolonged symptom (e.g., ongoing tinnitus for months). | Do NOT assume you are the first provider they are seeing. You MUST explicitly ask, "Have you seen an ENT, urgent care, or any other doctor for this yet?" before assigning tests. | Prevents missing critical prior diagnoses (e.g., TTTS) and prior imaging (e.g., MRI) that alter the diagnostic landscape. |

| ID | Section | Trigger | Action | Expected effect |
|---|---|---|---|---|
| 0056 | common_mistakes | The conversation initiates and the patient provides their opening statement. | Acknowledge the statement and ask an open-to-focused follow-up question. Do not halt without producing a message. | Ensures the patient receives an initial response and begins the consultation successfully. |
| 0122 | strategy | You have assigned multiple routine tests (e.g., PTA, Speech, Tymp) but the patient only uploads the first one. | Briefly interpret the first test and immediately prompt the patient to upload the next test in the sequence (e.g., "PTA shows profound loss. Please now upload the speech audiometry we discussed"). Do NOT ask "Does this make sense?" or wait for unnecessary agreement between routine tests. | Conserves conversational turns and ensures the full test battery is collected before the consultation times out. |
| 0130 | strategy | You have completed 2 turns of history gathering, OR you are running low on turns. | You MUST immediately transition to assigning baseline tests (e.g., PTA, Speech) or outputting the final structured diagnosis schema. Do NOT use more than 2 turns for subjective history-taking. | Prevents the consultation from timing out prematurely by forcing a transition to testing and diagnosis. |

Snapshot note: all 19 entries belong to the core layer in the frozen playbook; there are no staging or backlog entries in this evaluation-time snapshot. Stable IDs are retained solely to map each displayed rule to the unchanged YAML records; identifiers are assigned sequentially at backlog ingestion, and candidates that were not retained in the frozen core are not shown, so the displayed IDs are non-contiguous. The combined evaluator–reflector/curator lifecycle is described in Supplementary Note 4.

Section taxonomy: tactics encode short behavioural micro-rules; strategy encodes broader workflow priors; and common_mistakes encode failure modes to avoid.

## Supplementary Note 5. Component ablation across two backbones

This Note details the component-ablation analyses summarised in main-text §"Rubric-guided playbook augmentation is the largest contributor to the gain" (Results, Fig. 6) and the cross-backbone analyses summarised in Supplementary Figs. S3 and S4. The main analysis is the GPT-5 component ablation (Section 5.1), which underlies the rubric-level claims reported in the main text. An exploratory cross-backbone analysis on a Qwen-family backbone (Section 5.2) examined whether the component pattern observed with GPT-5 is specific to a single frontier backbone or also emerges when the same rubric-guided playbook procedure is applied to a backbone with lower baseline performance. Both analyses use the automated 21-item rubric (Supplementary Note 3) on the same 75-case curated evaluation pool (40 Chinese, 35 English).

### 5.1 GPT-5 component ablation

We evaluated the GPT-5 backbone under four single-component augmentations (Vanilla; Vanilla + Tools; Vanilla + RAG; Vanilla + Playbook) and the full integrated system, using the automated 21-item rubric on the 75-case curated evaluation pool (40 Chinese, 35 English; one case in Vanilla + RAG returned no parseable scores, yielding n = 74 for that configuration). The aim was to localise the source of the AI audiologist's rubric-level gain across components.

#### Ablation configurations

- Vanilla — GPT-5 backbone alone, with the consultation orchestration prompt but no playbook, no audiometric/image tools, and no retrieval. This is the minimal generalist baseline.
- Vanilla + Tools — Vanilla plus the multimodal interpretation tools for audiograms, tympanograms, speech-in-noise reports and related materials (Methods, §AI audiologist system architecture).
- Vanilla + RAG — Vanilla plus the retrieval-augmented grounding component over the audiology reference corpus (Methods, §AI audiologist system architecture).
- Vanilla + Playbook — Vanilla plus the same frozen 19-rule GPT-5 playbook used in the full system and the blinded human comparison (Methods, §Rubric-guided closed-loop engineering; Supplementary Table S3). The policy was loaded before inference and was not modified during evaluation.
- Full integrated system — all components combined (playbook + tools + RAG), corresponding to the AI audiologist evaluated against human audiologists in the main blinded comparison (Results, Fig. 4).

#### Evaluation protocol

Each configuration completed one consultation per case with the AI patient simulator (Supplementary Note 2). Resulting transcripts were scored on the 21-item rubric by the automated evaluator (Supplementary Note 3). Frozen run-level summaries over the n = 75 evaluation pool are reported below; Vanilla + RAG includes n = 74 because one Chinese case was absent after a parser failure. The recovered per-case file contains 374 GPT-5 and 375 Qwen case–configuration records. Five GPT-5 records each lack one dimension composite (three F1 and two F2 values); these missing values are retained as blanks and were not imputed. The original run-level summaries remain the source for the displayed configuration tables and main-text Figure 6, whereas Supplementary Fig. S4 displays the available per-case values. Owing to summary rounding and the use of available-case denominators, direct per-case recomputation can differ from the reported all-language summary by less than 0.004 points. The ablation is interpreted descriptively, and no additional retrospective inferential comparisons were introduced.

### Dimension-level results

Mean scores by configuration on the four primary dimensions (F1 history-taking, F2 patient-centred communication, F3-L diagnosis & management computed as the mean of M1 and M3 with M2 reported separately on its 1–4 scale, and F3-B safety-item accuracy computed as the proportion of M4–M10 items marked Yes). N-weighted aggregates are reported; CN/EN means are shown in Supplementary Fig. S3. This ablation-specific F3-L (M1, M3) differs from the F3-Likert composite used in the blinded human evaluation (Fig. 4 and Table 2), which also includes M2 multiplied by 1.25.

| Configuration | n | F1 | F2 | F3-L (M1, M3) | F3-B (acc.) | M2 (1–4) |
|---|---|---|---|---|---|---|
| Vanilla | 75 | 3.541 | 2.116 | 3.686 | 0.848 | 3.040 |
| Vanilla + Tools | 75 | 3.745 | 2.319 | 3.873 | 0.857 | 3.133 |
| Vanilla + RAG | 74 | 3.445 | 2.133 | 3.696 | 0.817 | 3.068 |
| Vanilla + Playbook | 75 | 4.192 | 3.051 | 3.880 | 0.863 | 3.173 |
| Full integrated system | 75 | 4.147 | 3.011 | 3.874 | 0.871 | 3.093 |

### Δ vs Vanilla (n-weighted)

Mean differences from the Vanilla baseline on each dimension. Larger Δ indicates greater improvement over the minimal backbone; negative Δ indicates regression.

| Comparison | ΔF1 | ΔF2 | ΔF3-L (M1, M3) | ΔF3-B | ΔM2 |
|---|---|---|---|---|---|
| Vanilla + Tools − Vanilla | +0.204 | +0.203 | +0.187 | +0.010 | +0.093 |
| Vanilla + RAG − Vanilla | −0.095 | +0.017 | +0.010 | −0.031 | +0.028 |
| Vanilla + Playbook − Vanilla | +0.651 | +0.936 | +0.194 | +0.015 | +0.134 |
| Full integrated system − Vanilla | +0.606 | +0.895 | +0.187 | +0.023 | +0.054 |

**Four observations follow directly from the Δ table:**

- Playbook is the largest single-component contributor on history-taking and communication. Vanilla + Playbook produces +0.65 on F1 and +0.94 on F2 relative to Vanilla, both substantially larger than the next-best single component (Tools at +0.20 / +0.20).
- Tools alone yield modest, selective gains, plausibly attributable to audiogram-grounded evidence supporting information provision and shared decision-making within the F2 axis. The effect is real but smaller than the playbook effect.
- RAG alone produces no overall gain (ΔF1 −0.10, ΔF2 +0.02) and a small decline on the F1 axis. The pattern is consistent with retrieved generic domain content interfering with patient-centred elicitation when not coupled with behavioural scaffolding.
- Adding tools and RAG to a playbook-equipped backbone (Full integrated system) does not produce additional rubric-level gains beyond Vanilla + Playbook on F1 or F2: ΔF1 vs Vanilla + Playbook is −0.05, ΔF2 is −0.04 (n-weighted; small differences in either direction are within the cross-language scatter of the means).

### Item-level results (Likert items only)

Per-item Δ vs Vanilla on the 13 Likert/ordinal items H1–H5, C1–C6, M1, M3 (M2 ordinal is reported above on the original 1–4 scale; binary items M4–M10 are summarised by F3-B accuracy in the dimension table). The pattern of Playbook gains concentrates mainly on the items where Vanilla GPT-5 is not already near ceiling.

| Item | Vanilla | + Tools | + RAG | + Playbook | Full system |
|---|---|---|---|---|---|
| H1 | 2.903 | 2.943 | 2.681 | 3.973 | 3.867 |
| H2 | 3.281 | 3.605 | 2.861 | 3.560 | 3.333 |
| H3 | 4.539 | 4.769 | 4.571 | 4.653 | 4.747 |
| H4 | 4.579 | 4.605 | 4.642 | 4.573 | 4.560 |
| H5 | 2.402 | 2.805 | 2.472 | 4.200 | 4.227 |
| C1 | 2.547 | 2.742 | 2.554 | 3.534 | 3.528 |
| C2 | 2.133 | 2.213 | 2.027 | 3.413 | 3.445 |
| C3 | 1.360 | 1.755 | 1.487 | 2.453 | 2.378 |
| C4 | 1.733 | 2.021 | 1.811 | 2.533 | 2.374 |
| C5 | 1.933 | 2.013 | 1.852 | 3.146 | 3.175 |
| C6 | 2.987 | 3.168 | 3.068 | 3.227 | 3.168 |
| M1 | 3.827 | 4.080 | 3.919 | 4.120 | 4.147 |
| M3 | 3.547 | 3.667 | 3.473 | 3.640 | 3.600 |

The defining ablation pattern is visible in this table:

- H3 (test interpretation) and H4 (clinical reasoning) are already near ceiling under Vanilla (4.54 and 4.58 respectively, n-weighted) and gain little from any augmentation — these are items the GPT backbone alone handles well.
- H1 (problem elicitation) and H5 (ideas, concerns, expectations and functional impact) are where Vanilla underperforms and where Playbook contributes the largest gains (H1: 2.90 → 3.97, +1.07; H5: 2.40 → 4.20, +1.80).
- F2 items C1–C5 gain substantially under Playbook, with C2 (+1.28), C5 (+1.21) and C3 (+1.09) the largest. Tools alone produce smaller selective gains on C3 and C4, while RAG alone produces near-zero or negative effects across F2.

### Interpretation

On F1 and F2, the ordering of components by contribution is Playbook ≫ Tools > RAG. The pattern is consistent with the central conceptual claim of the manuscript — that the bottleneck for specialist consultation quality in low-data medical domains is workflow policy rather than declarative knowledge — because:

- declarative knowledge is supplied either implicitly by the GPT backbone (general-purpose models already exceed 75% on audiology qualification examinations; Introduction, refs 13 and 14) or explicitly by RAG, and neither was the largest component of gain;
- in-context behavioural scaffolding via the playbook — encoding when to probe, how to organise the consultation, and what failure modes to avoid — accounts for the bulk of the measurable gain on the rubric, particularly on items dependent on consultation structure and patient-centred fluency.

## 5.2 Exploratory cross-backbone analysis on a Qwen-family backbone

To assess whether the component pattern observed with GPT-5 is specific to a single frontier backbone, we repeated the five-configuration automated-evaluator ablation on a Qwen-family backbone (Qwen3.5-Plus, accessed through the Alibaba Cloud DashScope OpenAI-compatible API). The patient simulator (GPT-5-mini), automated evaluator (Gemini 3.1 Pro Preview, Google Vertex AI), retrieval pipeline (RAGFlow; maximum five retrievals per turn), case set (40 Chinese and 35 English; n = 75 in every configuration), per-consultation turn cap and rubric (Supplementary Note 3) were held constant; only the audiologist backbone differed. This analysis used the automated rubric and was not part of the blinded human evaluation. It is therefore reported as exploratory evidence about the cross-backbone behaviour of the induction architecture rather than as an additional primary endpoint. Qwen configuration-level summaries are reported in the tables below; no additional retrospective paired significance tests were introduced.

### Setup and Qwen-specific differences from the GPT-5 ablation

Two methodological aspects of the Qwen ablation differ from the GPT-5 ablation and should be made explicit.

First, the Vanilla + Playbook and Full integrated system configurations on the Qwen backbone used a 7-rule playbook (playbook_v2_seed_qwen.yaml; available to editors and reviewers under confidential review) independently induced on Qwen under the same rubric-guided evaluator–reflector–curator procedure. The GPT-5 and Qwen analyses therefore test the procedure on separately induced, backbone-specific policies (19 and 7 rules, respectively), rather than transferring one fixed playbook between models.

Second, the Qwen runs used reasoning/thinking mode disabled (provider parameter ACE_DISABLE_THINKING=1; output mode set to native), reflecting the configuration under which the Qwen induction trajectory itself was conducted. This differs from the GPT-5 ablation, which used GPT-5 with reasoning_effort="medium". All other inference settings — patient simulator policy, automated evaluator prompts, retrieval defaults, audiogram/tympanogram tool implementations, otoscopy classifier, turn caps, retrieval-call caps, and sampling defaults — were held constant across the two backbones (Methods, §Inference settings and compute).

### Dimension-level results (Qwen)

Mean scores by configuration on the four primary dimensions for the Qwen backbone (n-weighted aggregates across CN + EN; n = 75 in every configuration). For comparability with Section 5.1, the same dimension definitions and the same n-weighting convention are used.

| Configuration | n | F1 | F2 | F3-L (M1, M3) | F3-B (acc.) | M2 (1–4) |
|---|---|---|---|---|---|---|
| Vanilla | 75 | 2.704 | 2.091 | 3.100 | 0.672 | 2.573 |
| Vanilla + Tools | 75 | 2.733 | 2.169 | 3.647 | 0.787 | 3.013 |
| Vanilla + RAG | 75 | 2.837 | 2.058 | 3.333 | 0.737 | 2.667 |
| Vanilla + Playbook | 75 | 3.360 | 2.656 | 3.207 | 0.676 | 2.627 |
| Full integrated system | 75 | 3.312 | 2.927 | 3.587 | 0.739 | 2.907 |

### Δ vs Vanilla (Qwen, n-weighted)

| Comparison | ΔF1 | ΔF2 | ΔF3-L (M1, M3) | ΔF3-B | ΔM2 |
|---|---|---|---|---|---|
| Vanilla + Tools − Vanilla | +0.029 | +0.078 | +0.547 | +0.114 | +0.440 |

| Comparison | ΔF1 | ΔF2 | ΔF3-L (M1, M3) | ΔF3-B | ΔM2 |
|---|---|---|---|---|---|
| Vanilla + RAG − Vanilla | +0.133 | −0.033 | +0.234 | +0.065 | +0.093 |
| Vanilla + Playbook − Vanilla | +0.656 | +0.565 | +0.107 | +0.004 | +0.053 |
| Full integrated system − Vanilla | +0.608 | +0.836 | +0.487 | +0.067 | +0.333 |

**Three observations follow from the Qwen Δ table:**

- The Qwen Vanilla baseline scores below the GPT-5 Vanilla baseline on F1 (2.70 vs 3.54), F3-L (3.10 vs 3.69) and F3-B (0.67 vs 0.85), with similar F2 (2.09 vs 2.12), indicating that Qwen3.5-Plus has a lower starting point than GPT-5 on most dimensions of this consultation task.
- Vanilla + Playbook produced the largest single-component gains on F1 (+0.66) and F2 (+0.57), the same two dimensions on which Playbook was the largest contributor for GPT-5 (Section 5.1). The F1 gain was directly comparable in magnitude to GPT-5 (+0.65); the F2 gain was smaller than GPT-5 (+0.94) but still substantial. Although the two playbooks were induced independently, the qualitative ordering of components — Playbook as the largest contributor on F1 and F2 — replicates across backbones.
- The Full integrated system showed a pattern qualitatively distinct from GPT-5: on Qwen, Full added substantial gains over Vanilla + Playbook alone on F2 (+0.27), F3-L (+0.38), and F3-B (+0.06), whereas on GPT-5 the Vanilla + Playbook and Full configurations were essentially indistinguishable on every dimension (Full − Playbook: F1 −0.05, F2 −0.04, F3-L −0.01, F3-B +0.01).

### Item-level results (Qwen, Likert items only)

Per-item Δ vs Vanilla on the 13 Likert/ordinal items H1–H5, C1–C6, M1, M3 for the Qwen backbone (n-weighted, n = 75 per configuration).

| Item | Vanilla | + Tools | + RAG | + Playbook | Full system |
|---|---|---|---|---|---|
| H1 | 1.706 | 1.507 | 2.026 | 2.854 | 2.507 |
| H2 | 1.466 | 1.320 | 1.747 | 2.427 | 2.026 |
| H3 | 4.440 | 4.800 | 4.347 | 4.440 | 4.760 |
| H4 | 4.173 | 4.627 | 4.120 | 4.373 | 4.640 |
| H5 | 1.733 | 1.414 | 1.947 | 2.707 | 2.627 |
| C1 | 2.840 | 2.707 | 2.773 | 3.800 | 4.013 |
| C2 | 1.853 | 1.574 | 1.920 | 2.400 | 2.320 |
| C3 | 2.600 | 2.774 | 2.293 | 2.627 | 3.000 |
| C4 | 1.254 | 1.173 | 1.347 | 1.306 | 1.520 |
| C5 | 2.173 | 2.160 | 2.040 | 3.787 | 3.853 |
| C6 | 1.827 | 2.627 | 1.973 | 2.013 | 2.854 |
| M1 | 3.400 | 4.013 | 3.573 | 3.533 | 3.867 |
| M3 | 2.800 | 3.280 | 3.093 | 2.880 | 3.307 |

The Qwen item-level pattern shows the same overall structural features as GPT-5: H3 (test interpretation, 4.44/5) and H4 (clinical reasoning, 4.17/5) are already near ceiling under Vanilla and gain little from the playbook, mirroring the GPT-5 pattern. H1 (problem elicitation: 1.71 → 2.85, +1.15) and H5 (ideas, concerns, expectations and functional impact: 1.73 → 2.71, +0.97) are where Vanilla underperforms and where Playbook contributes the largest gains, again paralleling the GPT-5 pattern in which the playbook drives most of the gain on structured patient-centred elicitation. Across the F2 communication items, the largest Playbook-driven gains on Qwen were on C5 responding to emotions (+1.61), C1 fostering the relationship (+0.96), and C2 gathering information (+0.55).

A Qwen-specific signature emerges on three items where Full exceeds Vanilla + Playbook by a substantial margin: C6 enabling behaviour with written planning (Vanilla 1.83 → Vanilla + Playbook 2.01 → Full 2.85; Full adds +0.84 over Vanilla + Playbook alone), M1 differential-diagnosis appropriateness (3.40 → 3.53 → 3.87; Full adds +0.33), and M3 management-plan appropriateness (2.80 → 2.88 → 3.31; Full adds +0.43). These items concentrate on concrete clinical content — written plans, named differentials and management recommendations — and are the dimensions on which audiogram/tympanogram tools and retrieval over the audiology reference corpus most plausibly contribute directly. The pattern is consistent with tools and retrieval supplying clinical content that the Qwen playbook alone does not fully encode, while leaving the playbook-mediated history-taking and relational-communication gains intact.

### Cross-backbone pattern

Two cross-backbone observations follow from the comparison of Section 5.1 (GPT-5) and Section 5.2 (Qwen):

**(i) The qualitative ordering of single-component contributions on F1 and F2 — Playbook as the largest contributor — replicates from GPT-5 to Qwen despite independently induced, backbone-specific playbooks. This is consistent with the rubric-guided playbook procedure producing useful consultation policies on backbones with different baseline performance; the result does not test direct transfer of a fixed rule set.**

**(ii)** The relative contribution of tools and retrieval beyond the playbook differs systematically between backbones. For GPT-5, Vanilla + Playbook and Full are essentially indistinguishable on every dimension (Full − Vanilla + Playbook: F1 −0.05, F2 −0.04, F3-L −0.01, F3-B +0.01). For Qwen, Full exceeds Vanilla + Playbook on three of the four dimensions (Full − Vanilla + Playbook: F1 −0.05, F2 +0.27, F3-L +0.38, F3-B +0.06), with the Full-system advantage concentrated on items dependent on concrete clinical content (C6 written planning, M1 differentials, M3 management plan; see Item-level results above). One plausible interpretation, based on a single alternative backbone, is that tools and retrieval are most valuable when the underlying backbone has less inherent capability on these content-driven dimensions, while the playbook-mediated lift on relational and structural dimensions is broadly comparable across backbones. This interpretation is exploratory; cross-backbone generalisation about the relative contribution of tools and retrieval would require evaluation on additional backbones spanning a wider capability range.

### Limitations of the cross-backbone analysis

Three limitations bound the interpretation of Section 5.2.

First, the Qwen analysis used the automated evaluator and was not part of the blinded human evaluation (Results, Fig. 4); it provides evidence about the structure of component contributions on a different backbone, not evidence of clinical consultation quality comparable to the blinded human comparison. The component-pattern claims in Section 5.2 are descriptive and were not subjected to retrospective case-level hypothesis testing.

Second, the Vanilla + Playbook and Full configurations on the Qwen backbone used a Qwen-induced 7-rule playbook, not the GPT-5-induced 19-rule playbook (Section Setup). The cross-backbone evidence therefore concerns the rubric-guided induction procedure and the policies it produces, rather than transfer of a fixed playbook; this distinction is preserved in the interpretation above.

Third, the analysis covers a single alternative backbone (Qwen3.5-Plus); whether the cross-backbone pattern would generalise to smaller open-weight models or to other frontier backbones cannot be inferred from these data and remains an open empirical question.

### Methods note on descriptive interpretation

The per-case analysis file retains all 749 case–configuration records, with five unavailable GPT-5 dimension composites represented explicitly as missing; these records underlie the available-case distribution displays in Supplementary Fig. S4. The frozen run-level summaries remain the source for the configuration-level tables; per-case records are available to editors and reviewers under confidential review. Because the ablation was exploratory, used an automated evaluator and was outside the blinded human-rated primary endpoint, we did not add retrospective paired P values, confidence intervals or effect-size claims across configurations. The primary AI–Human comparison (Fig. 4 and Table 2) used blinded human raters and was analysed inferentially as pre-specified.

## Figure references

Supplementary Fig. S3. Heatmap of mean evaluation scores across the five ablation configurations on the four primary dimensions, shown for both the GPT-5 backbone (Section 5.1) and the Qwen backbone (exploratory cross-backbone analysis, Section 5.2). Rows are configuration × language and columns are dimensions; cell colour is the within-metric normalised score (0 = column minimum, 1 = column maximum). Source: the configuration-level summaries reported in this Note.

Supplementary Fig. S4. Per-configuration distribution-stability panel for both backbones. The two provider rows (GPT-5 and Qwen) are shown across four dimension columns. Within each subplot, the available-case configuration mean ± 1 s.d., minimum–maximum range and individual case scores are displayed; missing dimension composites are omitted without imputation. The figure is descriptive and does not present retrospective paired inference. Source: per-case automated-evaluator records (available to editors and reviewers under confidential review) and the configuration-level summaries reported in this Note.

**Supplementary Fig. S3**

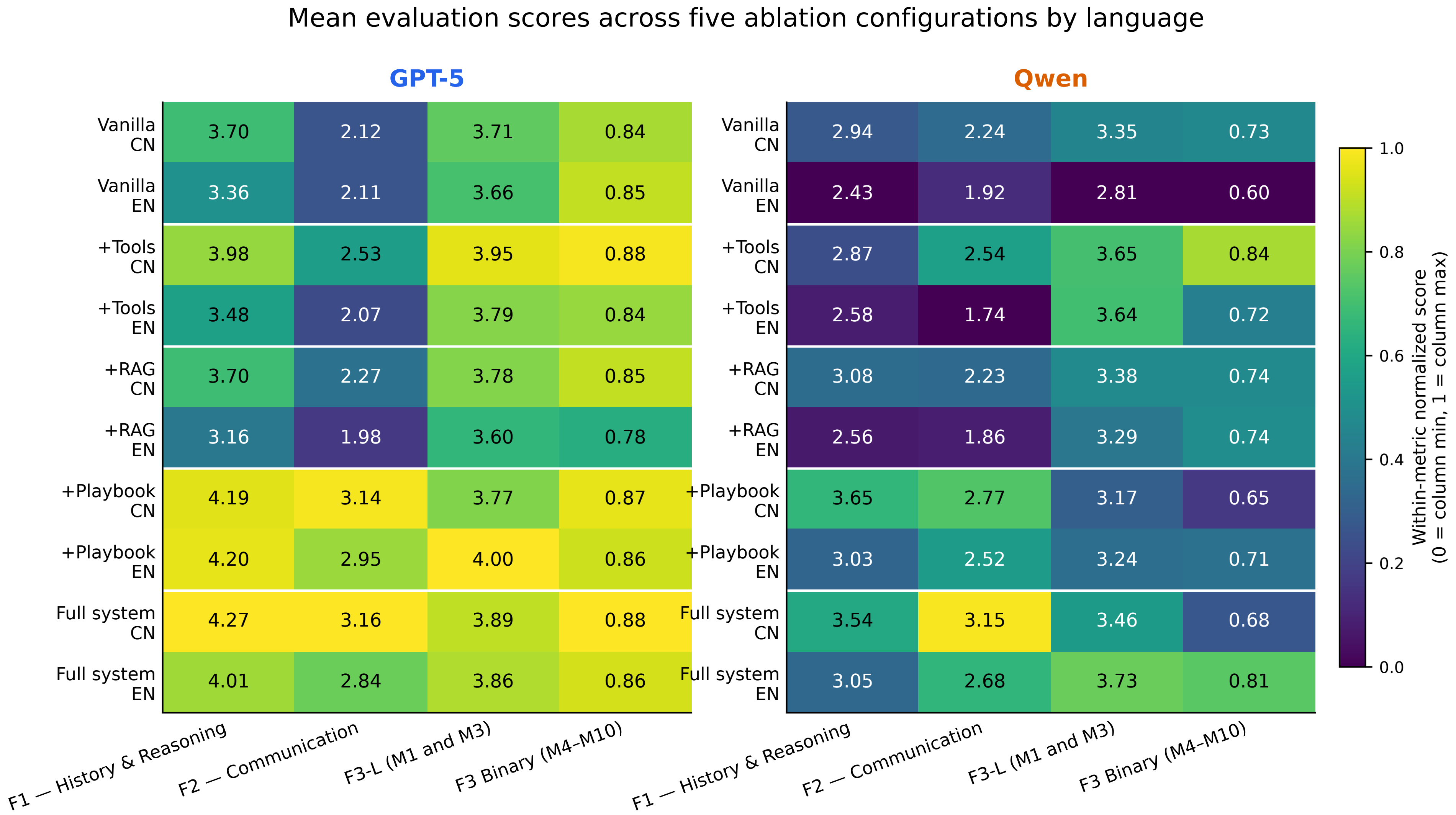

**Supplementary Fig. S4**

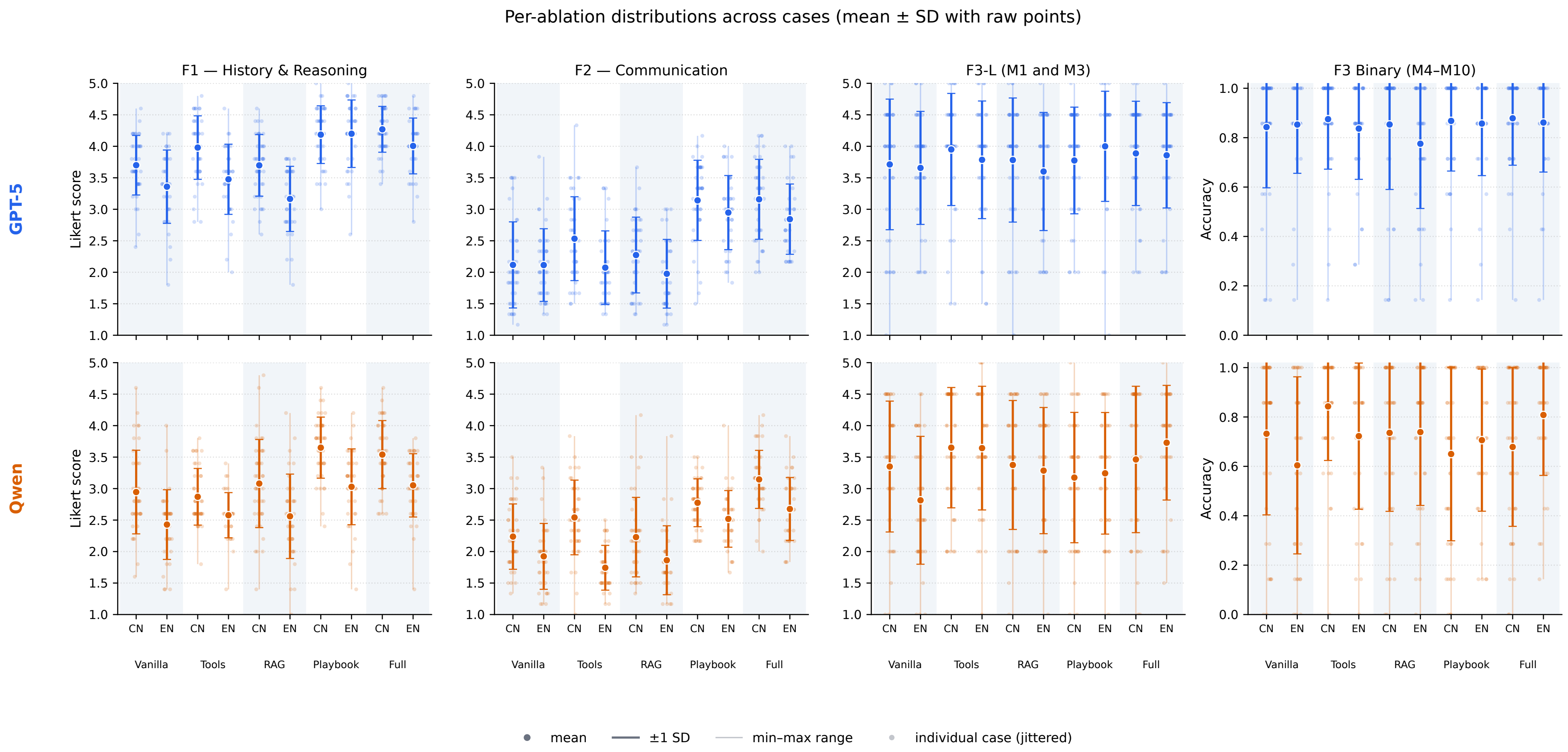

## Supplementary Note 6. Detailed human rating results

This Note expands the human-rating analysis reported in main-text §"The AI audiologist outperforms human audiologists in blinded evaluation" (Results, Fig. 4) and Methods §"Human rating protocol" with per-rater × per-dimension breakdowns and per-item × per-language paired differences. All numbers below are computed from the analysed set of 232 ratings on 58 cases (30 Chinese, 28 English) introduced in Methods.

### Per-rater × per-dimension paired Δ

Each rater scored AI and Human consultations for a subset of cases. For each rater, the table reports the paired AI − Human mean difference on the overall composite and each of the three Likert dimensions (F1 history-taking, F2 patient-centred communication, F3-Likert diagnosis & management), restricted to cases where that rater scored both AI and Human. Rater identity is masked behind R1–R4; R3 scored Chinese-language consultations only (Methods, §Human rating protocol).

| Rater | Dimension | n | AI mean | Human mean | Δ | 95% CI | d | P |
|---|---|---|---|---|---|---|---|---|
| R1 | Overall composite | 28 | 4.81 | 3.62 | +1.19 | [+0.85, +1.53] | 1.37 | $9.0\times10^{-8}$ |
| R1 | F1 history | 28 | 4.88 | 3.71 | +1.16 | [+0.78, +1.55] | 1.17 | $1.3\times10^{-6}$ |
| R1 | F2 communication | 28 | 4.85 | 3.47 | +1.38 | [+0.97, +1.79] | 1.30 | $2.2\times10^{-7}$ |
| R1 | F3-Likert | 28 | 4.71 | 3.69 | +1.02 | [+0.67, +1.37] | 1.13 | $2.2\times10^{-6}$ |
| R2 | Overall composite | 30 | 4.01 | 3.37 | +0.64 | [+0.45, +0.82] | 1.30 | $7.9\times10^{-8}$ |
| R2 | F1 history | 30 | 3.92 | 3.42 | +0.50 | [+0.32, +0.68] | 1.05 | $3.0\times10^{-6}$ |
| R2 | F2 communication | 30 | 3.91 | 3.19 | +0.72 | [+0.50, +0.93] | 1.23 | $2.1\times10^{-7}$ |
| R2 | F3-Likert | 30 | 4.20 | 3.50 | +0.70 | [+0.49, +0.91] | 1.23 | $2.1\times10^{-7}$ |
| R3 | Overall composite | 26 | 4.77 | 2.43 | +2.34 | [+2.05, +2.64] | 3.23 | $6.3\times10^{-15}$ |
| R3 | F1 history | 26 | 4.85 | 2.43 | +2.42 | [+2.07, +2.78] | 2.77 | $2.0\times10^{-13}$ |
| R3 | F2 communication | 26 | 4.90 | 2.37 | +2.53 | [+2.16, +2.91] | 2.73 | $2.7\times10^{-13}$ |
| R3 | F3-Likert | 26 | 4.57 | 2.49 | +2.07 | [+1.70, +2.45] | 2.24 | $2.0\times10^{-11}$ |
| R4 | Overall composite | 31 | 3.74 | 2.37 | +1.38 | [+1.14, +1.61] | 2.14 | $6.5\times10^{-13}$ |
| R4 | F1 history | 31 | 3.83 | 2.79 | +1.05 | [+0.68, +1.41] | 1.06 | $1.7\times10^{-6}$ |
| R4 | F2 communication | 31 | 3.41 | 1.78 | +1.63 | [+1.36, +1.90] | 2.23 | $2.3\times10^{-13}$ |
| R4 | F3-Likert | 31 | 3.99 | 2.53 | +1.45 | [+1.03, +1.87] | 1.27 | $7.2\times10^{-8}$ |

### Per-item paired Δ across all 21 rubric items

Item-level paired AI − Human differences (case-level means averaged across the two raters per consultation, n = 58 cases). q values use Benjamini–Hochberg false-discovery-rate correction across the 21 item-level tests; the only item without significant AI advantage is M8 (absence of confabulation; Δ = +0.03, q = 0.47); this non-significant difference should not be interpreted as equivalence. Binary items M4–M10 are scored on a 0/1 scale.

| Item | AI mean | Human mean | Δ | 95% CI | d | P | q (BH-FDR) |
|---|---|---|---|---|---|---|---|
| H1 Problem elicitation | 4.362 | 3.466 | +0.897 | [+0.656, +1.138] | 0.98 | $5.7\times10^{-10}$ | $9.1\times10^{-10}$ |
| H2 Audiology history | 4.224 | 3.198 | +1.026 | [+0.744, +1.308] | 0.96 | $1.1\times10^{-9}$ | $1.7\times10^{-9}$ |
| H3 Test interpretation | 4.319 | 2.793 | +1.526 | [+1.239, +1.812] | 1.40 | $3.4\times10^{-15}$ | $8.9\times10^{-15}$ |
| H4 Clinical reasoning | 4.353 | 3.060 | +1.293 | [+1.066, +1.520] | 1.50 | $2.5\times10^{-16}$ | $1.0\times10^{-15}$ |
| H5 ICE & impact | 4.431 | 3.009 | +1.422 | [+1.133, +1.711] | 1.29 | $6.5\times10^{-14}$ | $1.4\times10^{-13}$ |
| C1 Fostering relationship | 4.310 | 3.009 | +1.302 | [+1.054, +1.549] | 1.38 | $5.4\times10^{-15}$ | $1.3\times10^{-14}$ |
| C2 Gathering information | 4.293 | 2.957 | +1.336 | [+1.097, +1.575] | 1.47 | $5.3\times10^{-16}$ | $1.8\times10^{-15}$ |
| C3 Providing information | 4.138 | 2.776 | +1.362 | [+1.117, +1.607] | 1.46 | $6.1\times10^{-16}$ | $1.8\times10^{-15}$ |
| C4 Shared decision-making | 4.155 | 2.431 | +1.724 | [+1.475, +1.973] | 1.82 | $6.4\times10^{-20}$ | $6.8\times10^{-19}$ |
| C5 Responding to emotions | 4.241 | 2.698 | +1.543 | [+1.280, +1.806] | 1.54 | $7.7\times10^{-17}$ | $4.0\times10^{-16}$ |
| C6 Enabling behaviour | 4.250 | 2.328 | +1.922 | [+1.685, +2.160] | 2.13 | $5.0\times10^{-23}$ | $1.0\times10^{-21}$ |
| M1 Differential diagnosis appropriateness | 4.198 | 3.112 | +1.086 | [+0.854, +1.319] | 1.23 | $4.1\times10^{-13}$ | $7.1\times10^{-13}$ |
| M2 Differential diagnosis comprehensiveness | 4.537 | 2.974 | +1.562 | [+1.304, +1.821] | 1.59 | $2.0\times10^{-17}$ | $1.4\times10^{-16}$ |
| M3 Management plan appropriateness | 4.293 | 3.112 | +1.181 | [+0.934, +1.428] | 1.26 | $1.8\times10^{-13}$ | $3.5\times10^{-13}$ |
| M4 Appropriate investigations recommended | 0.931 | 0.776 | +0.155 | [+0.073, +0.238] | 0.49 | $3.9\times10^{-4}$ | $4.6\times10^{-4}$ |
| M5 Inappropriate investigations avoided | 0.966 | 0.862 | +0.103 | [+0.027, +0.180] | 0.35 | 0.009 | 0.0098 |
| M6 Appropriate treatments/counselling | 0.871 | 0.647 | +0.224 | [+0.128, +0.320] | 0.61 | $1.8\times10^{-5}$ | $2.5\times10^{-5}$ |
| M7 Inappropriate treatments avoided | 0.974 | 0.853 | +0.121 | [+0.046, +0.196] | 0.42 | 0.002 | 0.002 |
| M8 Confabulation absent | 0.940 | 0.914 | +0.026 | [-0.046, +0.097] | 0.10 | 0.472 | 0.472 |
| M9 Follow-up appropriate | 0.991 | 0.802 | +0.190 | [+0.095, +0.284] | 0.53 | $1.8\times10^{-4}$ | $2.2\times10^{-4}$ |
| M10 Escalation recommendation appropriate | 0.784 | 0.534 | +0.250 | [+0.131, +0.369] | 0.55 | $9.0\times10^{-5}$ | $1.2\times10^{-4}$ |

## Per-item × per-language paired Δ

Per-item paired AI − Human differences split by case language (CN n = 30, EN n = 28). The right-most column reports the CN − EN gap in paired Δ — i.e., how much larger the AI advantage is on Chinese cases than on English cases — and a Welch t-test on the per-case Δ distributions between languages. Items are sorted by descending CN − EN gap. The five items with the largest gap (H5, M3, H2, H3, C1) align with the items the main text highlights as bearing the cross-language asymmetry (Results, §Language stratification). Visual: Supplementary Fig. S5.

| Item | CN Δ | EN Δ | CN − EN gap | P_gap |
|---|---|---|---|---|
| H5 ICE & impact | +2.000 | +0.804 | +1.196 | $8.1\times10^{-6}$ |
| M3 Management plan appropriateness | +1.700 | +0.625 | +1.075 | $2.7\times10^{-6}$ |
| H2 Audiology history | +1.533 | +0.482 | +1.051 | $8.0\times10^{-5}$ |
| H3 Test interpretation | +2.017 | +1.000 | +1.017 | $2.3\times10^{-4}$ |
| C1 Fostering relationship | +1.750 | +0.821 | +0.929 | $6.7\times10^{-5}$ |
| C2 Gathering information | +1.767 | +0.875 | +0.892 | $7.5\times10^{-5}$ |

| Item | CN Δ | EN Δ | CN − EN gap | P_gap |
|---|---|---|---|---|
| C6 Enabling behaviour | +2.350 | +1.464 | +0.886 | $7.3\times10^{-5}$ |
| H1 Problem elicitation | +1.300 | +0.464 | +0.836 | $2.6\times10^{-4}$ |
| C5 Responding to emotions | +1.917 | +1.143 | +0.774 | 0.002 |
| C3 Providing information | +1.717 | +0.982 | +0.735 | 0.002 |
| H4 Clinical reasoning | +1.617 | +0.946 | +0.670 | 0.002 |
| M2 Differential diagnosis comprehensiveness | +1.875 | +1.228 | +0.647 | 0.011 |
| M1 Differential diagnosis appropriateness | +1.383 | +0.768 | +0.615 | 0.007 |
| C4 Shared decision-making | +2.000 | +1.429 | +0.571 | 0.019 |
| M9 Follow-up appropriate | +0.300 | +0.071 | +0.229 | 0.013 |
| M4 Appropriate investigations recommended | +0.250 | +0.054 | +0.196 | 0.015 |
| M7 Inappropriate treatments avoided | +0.200 | +0.036 | +0.164 | 0.026 |
| M6 Appropriate treatments/counselling | +0.283 | +0.161 | +0.123 | 0.201 |
| M10 Escalation recommendation appropriate | +0.283 | +0.214 | +0.069 | 0.562 |
| M5 Inappropriate investigations avoided | +0.100 | +0.107 | -0.007 | 0.926 |
| M8 Confabulation absent | +0.000 | +0.054 | -0.054 | 0.456 |

## Figure references

Fig. 4 (main). Primary blinded AI–Human comparison; paired-difference plot, dimension scores, item-level forest plot, language-stratified composite.

Supplementary Fig. S5. Per-item paired AI − Human differences stratified by language, with 95% CIs. Items are ordered by descending Chinese − English gap. Red circles and blue squares show the Chinese and English paired differences, respectively; horizontal error bars show the corresponding 95% CIs. Stars mark the five items with the largest Chinese − English gaps (H5, M3, H2, H3 and C1).

**Supplementary Fig. S5**

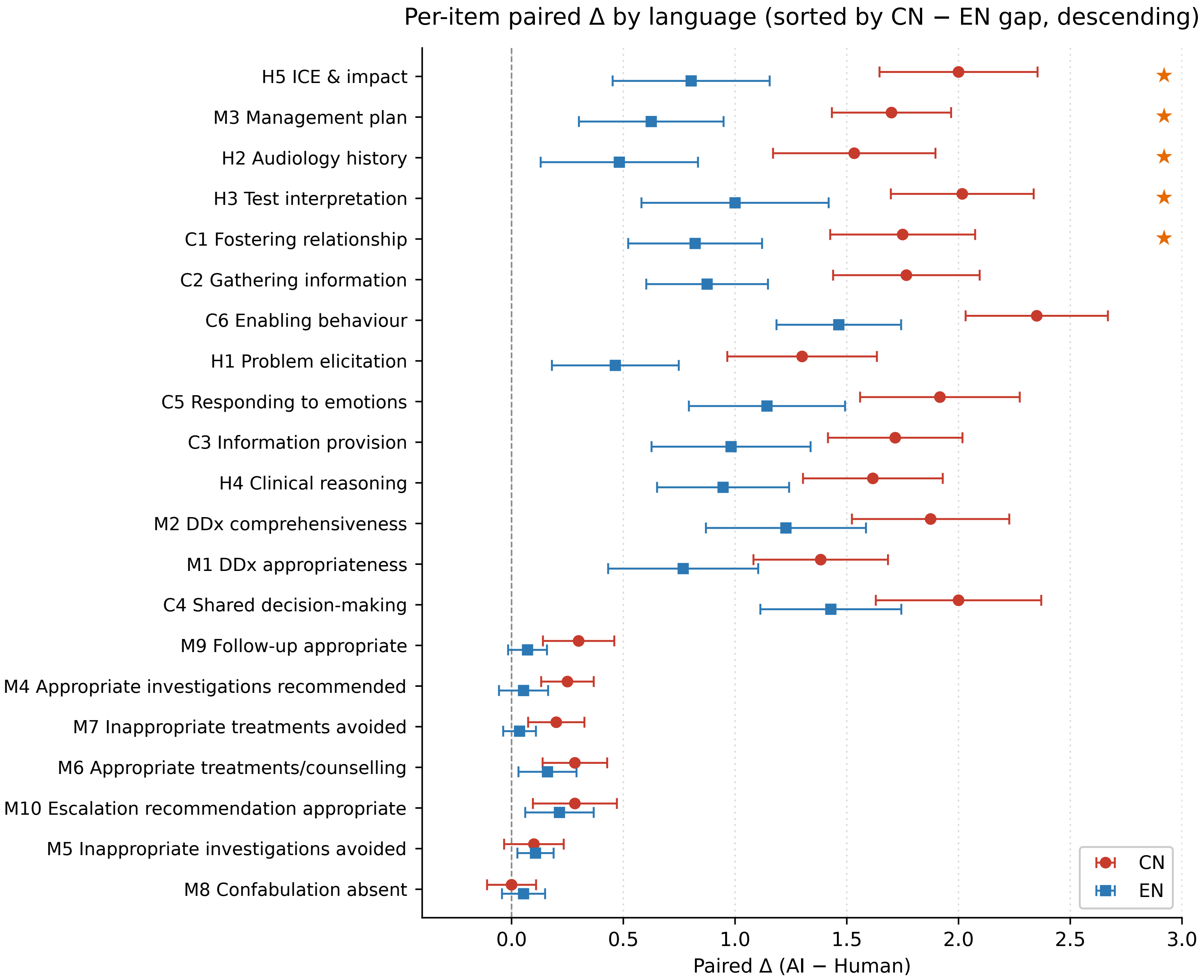

★ marks the five items with the largest CN − EN gap (H5, M3, H2, H3, C1).

## Supplementary Note 7. Inter-rater agreement and sensitivity analyses

### Overview

This Note reports inter-rater agreement statistics and rater-level sensitivity analyses for the primary blinded comparison (Results, §Rating agreement supports the robustness of the primary comparison; Fig. 5). The aims are to quantify agreement among the four blinded raters, assess whether the AI > Human conclusion depends on any single rater, and document the implementation of ICC(2,1) for the unbalanced two-rater-per-(case, source) design.

Design recap. Each (case, source) consultation was scored by exactly two of the four blinded audiologist raters (R1–R4); rater–case assignments were pre-specified so that each rater scored a balanced mix of AI and Human consultations across both languages within their assigned subset. Per-case rubric scores used in the primary analysis are the mean of the two raters who scored that (case, source). Raters were blinded to source identity (AI vs Human) and to one another's ratings. Throughout this Note rater labels are: R1, R2, R4 = bilingual raters; R3 = native-Chinese rater who scored Chinese-language consultations only.

### Inter-rater agreement (ICC and pairwise correlations)

Across the four primary scores, single-rater absolute-agreement ICC(2,1) values were moderate: 0.57 for the overall composite, 0.55 for F1, 0.53 for F2 and 0.54 for F3-Likert. Reliability of the two-rater mean was higher (ICC(2,k = 2) = 0.69–0.73). Pairwise Pearson correlations between rater pairs ranged from approximately r = 0.57 to 0.92 on the dimension scores (composite shown in Supplementary Fig. S6b). The higher Pearson correlations alongside moderate single-rater absolute-agreement ICCs indicate that raters ranked consultations more consistently than they matched one another's absolute score levels. The primary analysis therefore averaged the two ratings within each consultation and used case-matched AI–Human comparisons.

| Score | ICC(2,1) | ICC(2, k = 2) |
|---|---|---|
| **Overall composite (mean of F1, F2, F3-Likert)** | 0.57 | 0.73 |
| **F1 History-taking dimension** | 0.55 | 0.71 |
| **F2 Patient-centred communication** | 0.53 | 0.69 |
| **F3-Likert (Dx & Mgmt items)** | 0.54 | 0.70 |

**Item-level ICC.**

Across the 14 Likert/ordinal items (H1–H5, C1–C6 and M1–M3), ICC(2,1) ranged from 0.396 to 0.583 (median 0.45). The seven binary safety items (M4–M10) had lower ICC values (range 0.024–0.332; median 0.19). Several of these binary items had near-ceiling Yes rates, which restricted between-consultation variance and can reduce the stability of ICC estimates. For example, M9 (follow-up appropriate) had 83–100% pairwise agreement across rater pairs but ICC(2,1) = 0.33. Per-item ICC values are visualised in Supplementary Fig. S6a; pairwise agreement should be considered alongside ICC for these items.

| Item | ICC(2,1) |
| --- | --- |
| **H1 Problem elicitation** | 0.423 |
| **H2 Audiology history** | 0.398 |
| **H3 Test interpretation** | 0.396 |
| **H4 Clinical reasoning** | 0.438 |
| **H5 ICE & functional impact** | 0.547 |
| **C1 Fostering relationship** | 0.453 |
| **C2 Gathering information** | 0.467 |
| **C3 Information provision** | 0.411 |
| **C4 Shared decision-making** | 0.458 |
| **C5 Responding to emotions** | 0.488 |
| **C6 Enabling behaviour** | 0.513 |
| **M1 DDx appropriateness** | 0.400 |
| **M2 DDx comprehensiveness** | 0.583 |
| **M3 Management plan** | 0.398 |
| **M4 Appropriate investigations recommended** | 0.024 |
| **M5 Inappropriate investigations avoided** | 0.208 |
| **M6 Appropriate treatments/counselling** | 0.122 |
| **M7 Inappropriate treatments avoided** | 0.193 |
| **M8 Confabulation absent** | 0.103 |
| **M9 Follow-up appropriate** | 0.332 |
| **M10 Escalation recommendation appropriate** | 0.206 |

### Methods note on ICC implementation.

Because each (case, source) was scored by only two of the four raters, the design is not fully crossed and standard balanced-design ICC formulas (Shrout & Fleiss 1979) do not directly apply. We implemented ICC(2,1) absolute-agreement, single-rater, by computing the per-rater-pair ICC for each of the six possible rater pairs (subset to (case, source) units scored by both raters in the pair, $n \geq 16$ per pair) and aggregating across the six pairs using Fisher z-transformed weights proportional to $(n - 3)$. Aggregated ICCs reported above match those obtained from the equivalent crossed random-effects model (variance components: $\sigma^2$_target / ($\sigma^2$_target + $\sigma^2$_rater + $\sigma^2$_residual)) when that model converges. Implementation details are available to editors and reviewers under confidential review (see Code availability in the main manuscript).

## Per-rater paired Δ

All four raters had a positive mean AI − Human difference on the overall composite (table below). Per-rater paired differences ranged from +0.64 (R2) to +2.34 (R3), and all four per-rater paired tests had $P < 10^{-7}$. R3, who rated Chinese-language consultations only, had the largest mean difference; R2 had the smallest; R1 and R4 were intermediate. At the rater–case level, AI exceeded Human in 25 of 28 paired cases for R1, 25 of 30 for R2, 26 of 26 for R3 and 31 of 31 for R4. The remaining eight events comprised five Human-better events and three ties, as reported in the sign-test analysis below.

| Rater | n (paired) | Δ composite | 95% CI | Cohen's d | P (two-sided) |
|---|---|---|---|---|---|
| **R1** | 28 | +1.19 | [+0.85, +1.53] | 1.37 | $9.0\times10^{-8}$ |
| **R2** | 30 | +0.64 | [+0.45, +0.82] | 1.30 | $7.9\times10^{-8}$ |
| **R3** | 26 | +2.34 | [+2.05, +2.64] | 3.23 | $6.3\times10^{-15}$ |
| **R4** | 31 | +1.38 | [+1.14, +1.61] | 2.14 | $6.5\times10^{-13}$ |

**Interpretation.**

The between-rater range in mean paired differences was substantial, but even the smallest per-rater estimate (R2: Δ = +0.64, Cohen's d = 1.30) had a 95% CI entirely above zero. Variation in absolute score levels and effect magnitude across raters is consistent with the moderate, rather than excellent, absolute-agreement ICCs. The pooled case-matched analysis and the leave-one-rater-out analyses are therefore reported alongside the agreement estimates.

## Leave-one-rater-out sensitivity

To assess whether the primary result depended on any single rater, we recomputed the overall composite after excluding each rater in turn (table below; main-text Fig. 5c). The estimated AI − Human difference remained between +1.10 and +1.60, Cohen's d remained between 1.39 and 1.93, and all paired P values were below $10^{-14}$. Excluding R3 produced the smallest estimate (Δ = +1.10), whereas excluding R2 produced the largest (Δ = +1.60), consistent with their respective within-rater mean differences. Every leave-one-rater-out analysis preserved the direction and statistical conclusion of the primary result.

| Configuration | n | Δ composite | 95% CI | Cohen's d | P |
|---|---|---|---|---|---|
| **All four raters** | 58 | +1.35 | [+1.15, +1.54] | 1.84 | $4.5\times10^{-20}$ |
| **Drop R1** | 58 | +1.36 | [+1.16, +1.57] | 1.73 | $5.9\times10^{-19}$ |
| **Drop R2** | 58 | +1.60 | [+1.38, +1.82] | 1.93 | $4.9\times10^{-21}$ |
| **Drop R3** | 58 | +1.10 | [+0.93, +1.27] | 1.71 | $9.7\times10^{-19}$ |
| **Drop R4** | 58 | +1.32 | [+1.07, +1.57] | 1.39 | $4.8\times10^{-15}$ |

## Rater-level paired sign test

Pooling across all rater × case pairs ($n = 115$ rater–case scoring events in which the rater scored both the AI and the Human consultation for the same case), the AI exceeded the Human comparator on 107 of 115 events (93.0%), with 5 events where the Human exceeded the AI and 3 events tied. A one-sided sign test on the 112 non-tied events yields $P < 10^{-20}$, providing a non-parametric corroboration of the primary parametric result that does not depend on any distributional assumption.

## Figure references

Fig. 5 (main). Rater agreement and robustness. (a) ICC(2,1) and ICC(2,k = 2) by score. (b) Per-rater paired AI − Human differences; the displayed n values are paired-case counts (R1 28, R2 30, R3 26 and R4 31), not rating-record counts. (c) Leave-one-rater-out difference with the full-sample reference estimate.

Supplementary Fig. S6. Detailed rater agreement. (a) Per-item ICC(2,1) for all 21 rubric items. (b) Pairwise Pearson correlations for the overall composite ($r = 0.70$–$0.92$).

**Supplementary Fig. S6**

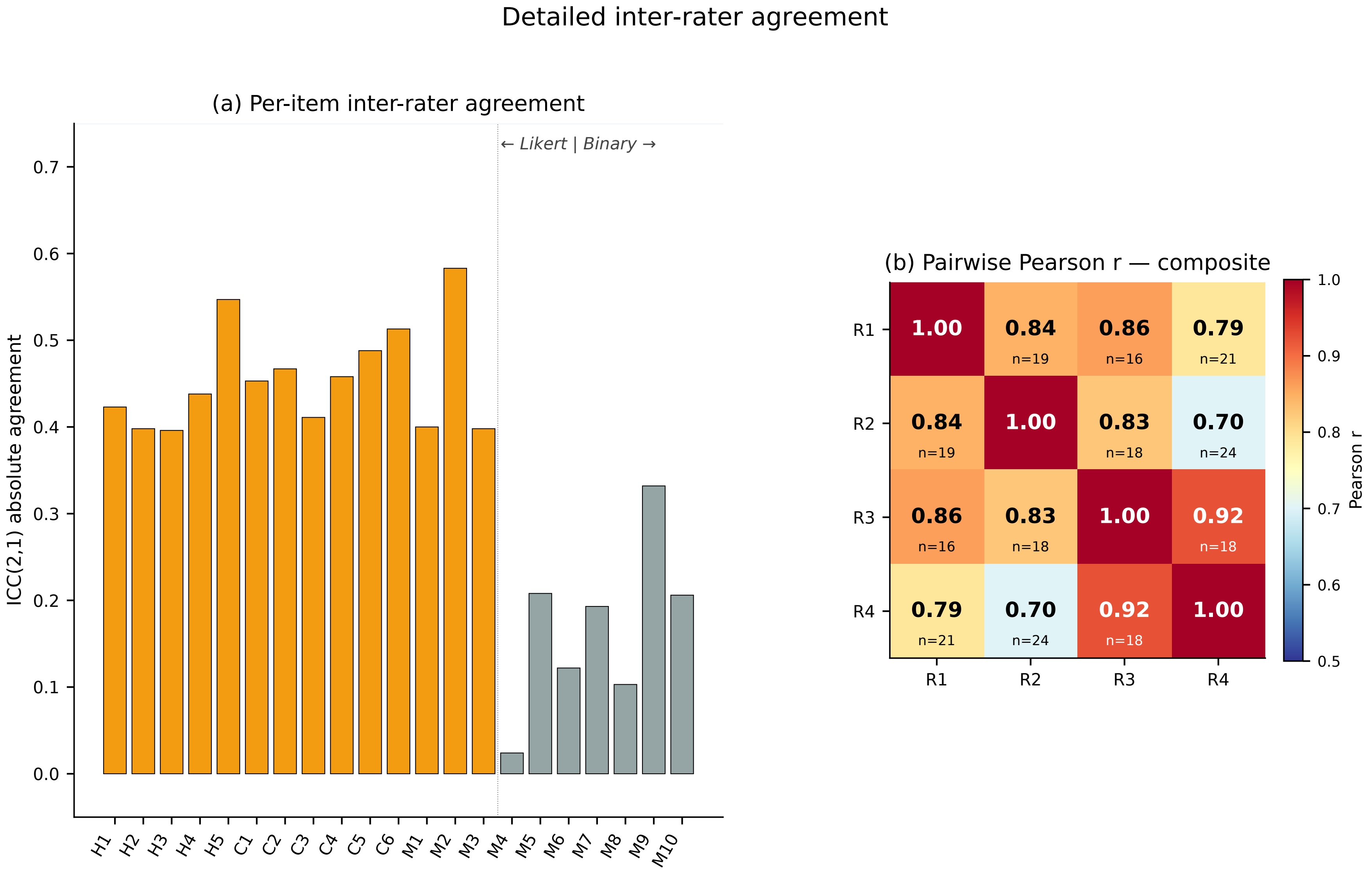

# Supplementary Note 8. Case-level low-score and comparison-event analysis

This Note provides a quantitative case-level analysis of the frozen 58-case blinded comparison. It reports low-score events, item-level Human-better events and ties. It does not infer transcript-level clinical failure modes because the score data alone do not support that level of qualitative interpretation.

## Operational definitions

For Likert and ordinal items, a low-score event was defined as a two-rater case mean ≤ 2.0. For binary items, a failure required both retained raters to mark No for the same case and item (case mean = 0); a mixed Yes/No pair was not classified as a failure. A Human-better event indicates that the Human case-level mean exceeded the AI case-level mean for the same item, and a tie indicates equal case-level means. These are item-level events and should not be interpreted as case-level composite wins or transcript-level diagnostic errors.

## Per-item case-level events

The table reports means and event counts for all 21 rubric items. M2 values were multiplied by 1.25 for inclusion in the composite.

| Item | Description | Type | AI mean | Human mean | AI low | Human low | Human-better | Tied |
|---|---|---|---|---|---|---|---|---|
| H1 | Problem elicitation and structure | Likert/ordinal | 4.362 | 3.466 | 0 | 3 | 5 | 14 |
| H2 | Audiology-specific history coverage | Likert/ordinal | 4.224 | 3.198 | 1 | 9 | 5 | 12 |
| H3 | Test interpretation | Likert/ordinal | 4.319 | 2.793 | 1 | 18 | 3 | 6 |
| H4 | Clinical judgment | Likert/ordinal | 4.353 | 3.060 | 0 | 11 | 1 | 5 |
| H5 | ICE and functional impact | Likert/ordinal | 4.431 | 3.009 | 0 | 11 | 1 | 11 |
| C1 | Fostering the relationship | Likert/ordinal | 4.310 | 3.009 | 0 | 9 | 1 | 7 |
| C2 | Gathering information | Likert/ordinal | 4.293 | 2.957 | 0 | 12 | 0 | 7 |
| C3 | Providing information clearly | Likert/ordinal | 4.138 | 2.776 | 0 | 16 | 3 | 7 |
| C4 | Shared decision-making | Likert/ordinal | 4.155 | 2.431 | 0 | 28 | 0 | 3 |
| C5 | Responding to emotions | Likert/ordinal | 4.241 | 2.698 | 0 | 22 | 1 | 6 |
| C6 | Enabling behavior | Likert/ordinal | 4.250 | 2.328 | 0 | 27 | 1 | 0 |
| M1 | Differential diagnosis appropriateness | Likert/ordinal | 4.198 | 3.112 | 0 | 13 | 4 | 5 |
| M2 | Differential-diagnosis comprehensiveness (multiplied by 1.25) | Likert/ordinal | 4.537 | 2.974 | 0 | 11 | 0 | 7 |
| M3 | Management plan appropriateness | Likert/ordinal | 4.293 | 3.112 | 0 | 10 | 4 | 8 |
| M4 | Appropriate investigations recommended | Binary | 0.931 | 0.776 | 1 | 2 | 3 | 36 |
| M5 | Inappropriate investigations avoided | Binary | 0.966 | 0.862 | 0 | 3 | 2 | 45 |
| M6 | Appropriate treatments/counselling | Binary | 0.871 | 0.647 | 0 | 7 | 3 | 31 |
| M7 | Inappropriate treatments avoided | Binary | 0.974 | 0.853 | 0 | 3 | 1 | 45 |
| M8 | Confabulation absent | Binary | 0.940 | 0.914 | 0 | 1 | 6 | 44 |
| M9 | Follow-up appropriate | Binary | 0.991 | 0.802 | 0 | 8 | 0 | 44 |
| M10 | Escalation recommendation appropriate | Binary | 0.784 | 0.534 | 1 | 17 | 6 | 27 |

## Descriptive findings

The AI had one low-score event on each of H2, H3, M4 and M10 and none on the remaining 17 items. The largest Human low-score counts were observed for C4 (28 cases), C6 (27), C5 (22), H3 (18), M10 (17) and C3 (16). These counts describe thresholded item scores and do not identify the clinical cause of an individual score.

Human-better events were most frequent for M8 and M10 (six cases each), followed by H1 and H2 (five cases each). These are item-level comparisons: the AI overall composite exceeded the Human composite in all 58 paired cases reported in the primary analysis. Ties were common on near-ceiling binary items.

Supplementary Fig. S7. Case-level low-score counts by rubric item. Blue and orange horizontal bars show the number of AI and Human cases, respectively, meeting the low-score definition. Items are ordered by descending Human low-score count. For Likert/ordinal items, the threshold is a two-rater case mean $\leq 2.0$; for binary items, failure requires both retained raters to mark No.

## Interpretive constraints

This analysis is descriptive and was conducted on the same frozen 58-case set as the primary comparison. Thresholded counts discard score magnitude and should not be used to diagnose transcript-level error mechanisms. Per-item event counts are reported in the table above; case-level event flags are available to editors and reviewers under confidential review, and consultation free text is not released.

## Supplementary Fig. S7

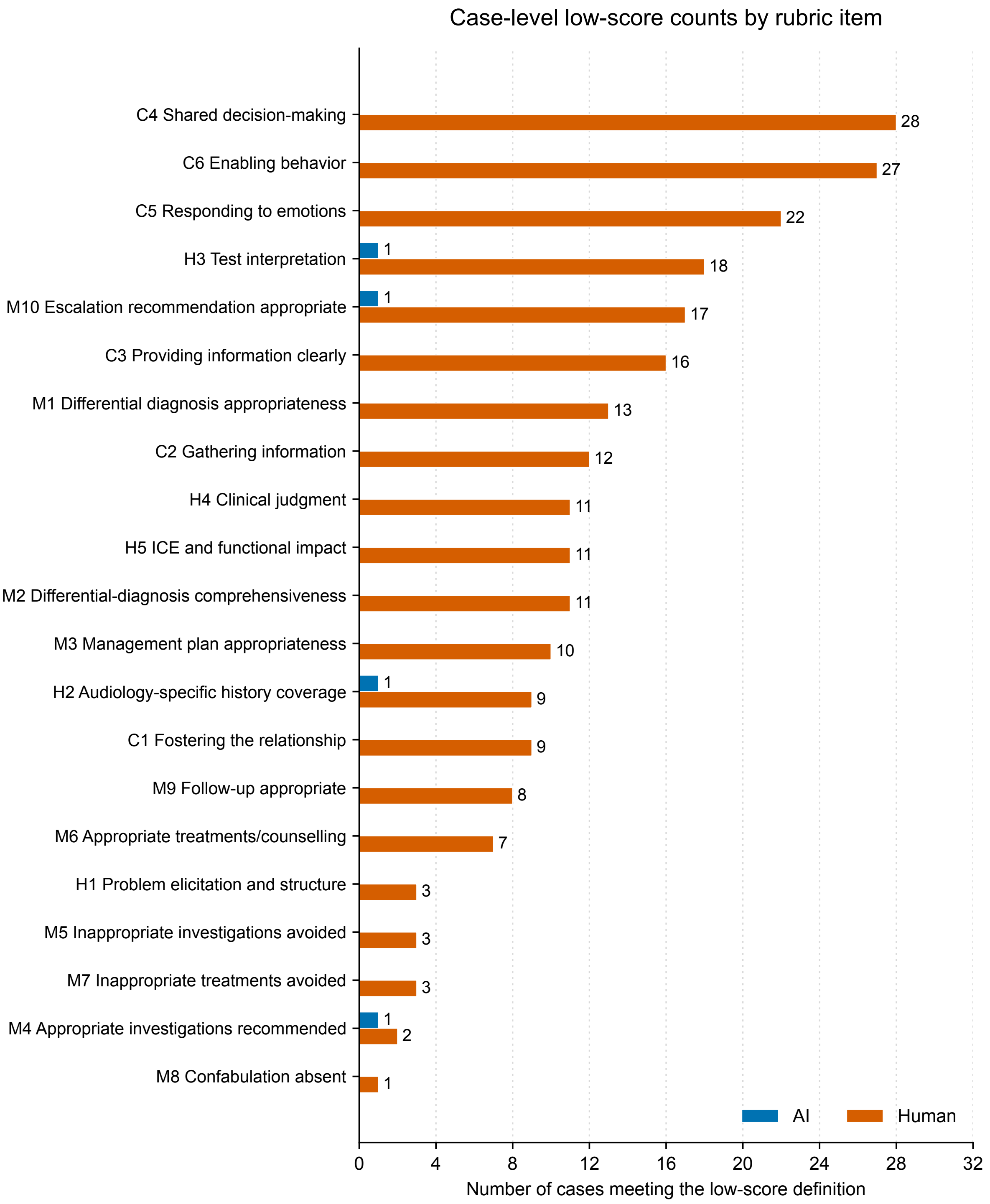